\documentclass[10pt,twocolumn,letterpaper]{article}

\usepackage{iccv}
\usepackage{times}
\usepackage{epsfig}
\usepackage{graphicx}
\usepackage{amsmath}
\usepackage{amssymb}
\usepackage{mdframed}
\usepackage{booktabs} 
\usepackage{subfigure}
\usepackage{microtype}
\usepackage{amsfonts}
\usepackage{rotating}
\usepackage{soul}

\usepackage[dvipsnames]{xcolor}
\definecolor{darkgreen}{rgb}{0.0, 0.5, 0.0}

\usepackage[breaklinks=true,bookmarks=false]{hyperref}

\iccvfinalcopy 

\def\iccvPaperID{7786} 

\ificcvfinal\fi

\begin{document}

\title{Membership Inference Attacks are Easier on Difficult Problems}

\author{Avital Shafran \qquad Shmuel Peleg \qquad Yedid Hoshen\\
The Hebrew University of Jerusalem, Israel\\
}

\maketitle
\ificcvfinal\thispagestyle{empty}\fi

\begin{abstract}
Membership inference attacks (MIA) try to detect if data samples were used to train a neural network model, e.g. to detect copyright abuses. 
We show that models with higher dimensional input and output are more vulnerable to MIA, and address in more detail models for image translation and semantic segmentation, including medical image segmentation.
We show that reconstruction-errors can lead to very effective MIA attacks as they are indicative of memorization. Unfortunately, reconstruction error alone is less effective at discriminating between non-predictable images used in training and easy to predict images that were never seen before. To overcome this, we propose using a novel predictability error that can be computed for each sample, and its computation does not require a training set. Our membership error, obtained by subtracting the predictability error from the reconstruction error, is shown to achieve high MIA accuracy on an extensive number of benchmarks. \footnote{Our source code is available at GitHub: https://bit.ly/3k0UE6P\\}
\end{abstract}
\vspace{-10pt}
\section{Introduction}

Deep neural networks have been widely adopted in various computer vision tasks, e.g.\ image classification, semantic segmentation, image translation and generation, etc. Due to the high sample-complexity of such models, they require large amounts of training data. However, 
collection and annotation of many training samples is an expensive and labor intensive process. In many domains, such as medical imaging, publicly available training data are particularly scarce due to privacy concerns. In such settings, 
a common solution is training the model privately and then providing black-box access to the trained model. However, even black-box access may leak sensitive information about the training data. 

\textit{Membership inference attacks (MIA)} are one way to detect such leakage. Given access to a data sample, an attacker attempts to find whether or not the sample was used in the training process. 

Due to memorization in deep neural networks, prediction confidence tends to be higher for images used in training. This difference in prediction confidence helps MIA methods to successfully determine which images were used for training. Therefore, in addition to detecting information leakage, MIA also provide insights on the degree of memorization in the victim model.

MIA has previously been applied to a variety of neural network tasks including: classification, generative adversarial models, and segmentation. The accuracy achieved by MIA can vary greatly as a function of different properties of the attempted tasks. Our empirical results highlight two properties that make tasks more vulnerable to MIA attacks: i) Uncertainty: tasks where there is more uncertainty in the prediction of the output given an input are more susceptible to MIA. ii) Output dimensionality: tasks with higher-dimensional outputs are more vulnerable to MIA. These apparently simple properties can explain non-intuitive observations, for example, that reconstruction-based MIA on CycleGAN leads to very low accuracy. 
As the training images should be reconstructed back after being translated to the other domain and back, there is no uncertainty of the output given the input.
As the desired output is fully specified by the input (i.e. they are identical), the reconstruction task is easy leading to low MIA accuracy.

Motivated by the above findings, we focus our attention on two tasks that exhibit these properties: supervised image translation and semantic segmentation, including medical image segmentation. 
We begin by evaluating a simple-to-implement but effective MIA that uses pixel-wise reconstruction error between the model output and ground truth. This approach exploits memorization of the training data in the victim model, resulting in lower reconstruction error on images used for training. However, we observe that for some sample images the ground truth result can be easily predicted, and for others it is harder to predict. Reconstruction error alone is therefore less accurate at discriminating between hard to predict samples used in training and easy samples not seen before. To overcome this limitation, we propose a novel predictability error which is computed for each query input image and its ground truth output.

Our predictability error uses the accuracy of a linear predictor computed over the given query image, predicting pixel values from deep features of the input image. The linear predictor serves as a simple approximation of the task attempted by the victim model, providing an indication of the ease of predicting the output image from the input. 
The reconstruction error, together with the predictability error, helps to discriminate between two factors of variation in the reconstruction error: (i) The "intrinsic" difficulty of the generation task for each image, based on its predictability error, and (ii) The boost in accuracy due to memorization of the training images. Defining a membership error that subtracts the predictability error from the reconstruction error is shown empirically to achieve high success rates in MIA. 

Differently from other MIA approaches, we do not assume the existence of a large number of in-distribution data samples for training a shadow model - but rather operate on a single sample, using only a single query to the victim model. Our method is demonstrated to be effective over strong baselines on an extensive number of benchmarks, taken from image translation, semantic segmentation, and medical image segmentation. 
We discuss possible defenses against MIA and show their ineffectiveness against our method. 

Our main contributions are:
\vspace{-6pt}
\begin{enumerate}
   \setlength{\itemsep}{1pt}
   \setlength{\parskip}{1pt}
   \setlength{\parsep}{1pt}
    \item Highlighting two key properties of tasks that are highly vulnerable to MIA. 
    \item Presenting the first MIA on image translation models.
    \item Proposing a general single-sample, self-supervised, image predictability error for MIA.
\end{enumerate}

\begin{table}[tb]
\centering
\small

\begin{tabular}{lccc}
\toprule
\textbf{Model} & \textbf{Dataset} & \textbf{Reconstruction} & \textbf{Membership} \\
\textbf{} & \textbf{} & \textbf{Error} & \textbf{Error} \\
\midrule
Pix2pix & Facades & 93.62\% & \textbf{97.59\%} \\
Pix2pix & Maps2sat & 84.22\% & \textbf{85.65\%} \\
Pix2pix & Cityscapes & 77.74\% & \textbf{83.23\%} \\
\midrule
Pix2pixHD & Facades & 98.92\% & \textbf{99.95\%} \\
Pix2pixHD & Maps2sat & 95.74\% & \textbf{99.42\%} \\
Pix2pixHD & Cityscapes & 96.04\% & \textbf{99.09\%} \\
\midrule
SPADE & Cityscapes & 99.75\% & \textbf{100\%}\\
SPADE & ADE20K & 85.31\% & \textbf{89.79\%}\\
\midrule
UperNet50 & ADE20K & 96.80\% & \textbf{98.09\%} \\
UperNet101 & ADE20K & 95.74\% & \textbf{96.94\%} \\
HRNetV2 & ADE20K & 83.67\% & \textbf{85.92\%} \\
\midrule
Inf-Net &  COVID19 & 97.16\% & \textbf{99.01\%} \\
PraNet & Polyp & 96.03\% & \textbf{96.38\%} \\
\bottomrule
\end{tabular}
\vspace{0.25cm}
\caption{Membership attack ROCAUC using our (i) reconstruction error $L_{rec}$ and (ii) membership error $L_{mem}$. Using the membership error, which subtracts the image predictability error from the reconstruction error, substantially improves performance.}

\label{tab:main_method_results}
\end{table}

\begin{figure}[t]
\begin{center}
\includegraphics[width=0.95\linewidth, height=0.45\linewidth]{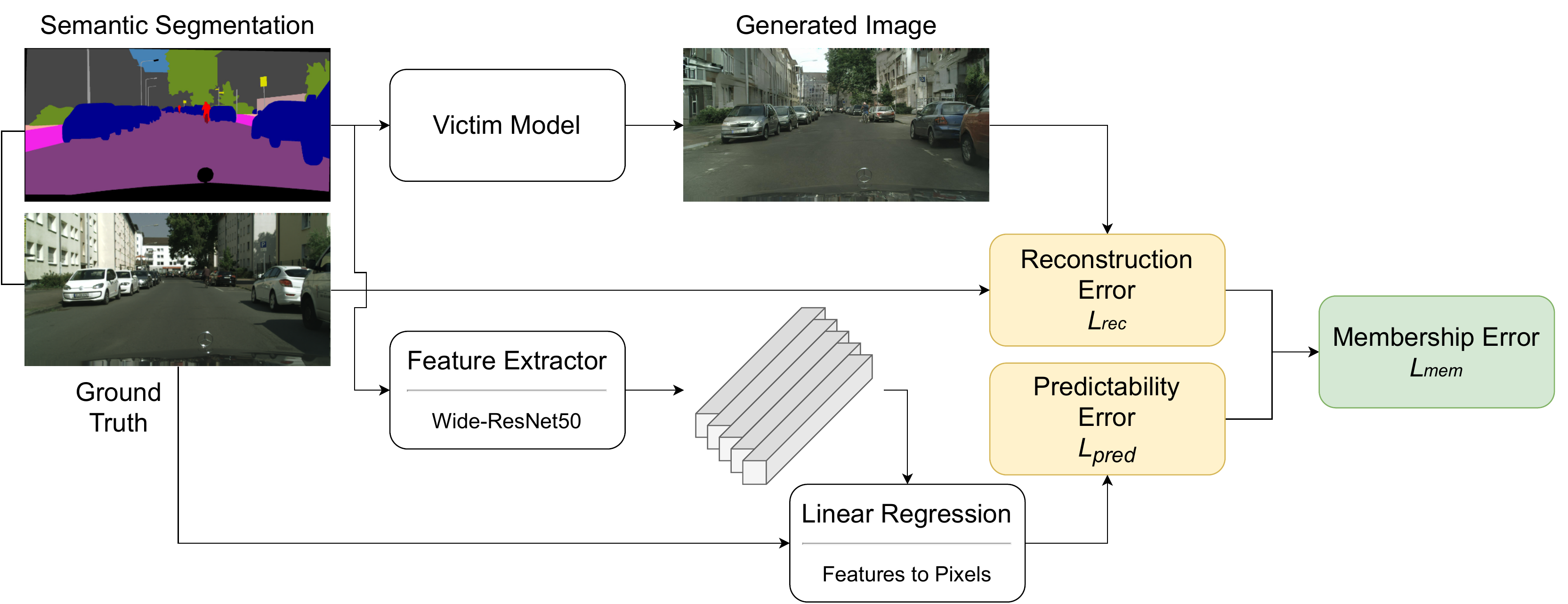}

\end{center}
\caption{Illustration of the proposed black-box membership inference attack. Here shown for the case of image translation over the Cityscapes dataset. We would like to determine if a given sample was used in training. The victim model predicts a reconstructed image based on the input. In the top path the difference between the reconstructed image and the ground truth image gives the reconstruction error $L_{rec}$. In the bottom path we compute the predictability error $L_{pred}$ of the sample from the error of a linear predictor to predict pixel values of the ground-truth image from deep features of the input. Subtracting $L_{pred}$ from $L_{rec}$ gives the membership error, $L_{mem}$.}
\label{fig:method_plot}
\end{figure}

\vspace{-10pt}
\section{Related Work}

\subsection{Membership Inference Attacks (MIA)}
Shokri et al. \cite{shokri2017membership} were the first to study MIA against classification models in a black-box setting. In this setting the attacker can only send queries to the victim model and get the full probability vector response, without being exposed to the model itself. They proposed to train multiple shadow models to mimic the behavior of the victim model, and then use those to train a binary classifier to distinguish between known samples from the train set and unknown samples. They assume the existence of in-distribution new training data and knowledge of the victim model architecture. 

Salem et al. \cite{salem2018ml} further relaxed those assumptions and demonstrated that using only one shadow model is sufficient, and proposed using out-of-distribution dataset and different shadow model architectures, for a slightly inferior attack. Even more interestingly, they showed that without any training, a simple  threshold on the victim model's confidence score is sufficient. This shows that classification models are more confident of samples that appeared in the training process, compared to unseen samples. 

Sablayrolles et al. \cite{sablayrolles2019white} proposed an attack based on applying a threshold over the loss value rather then the confidence and showed that black-box attacks are as good as white-box attacks. As the naive defense against such attacks is to modify the victim model's API to only output the predicted label, other works proposed label-only attacks \cite{yeom2018privacy, li2020label, choo2020label}.

While most previous work has been around classification models, there has been some effort regarding MIA on generative models such as GANs and VAEs \cite{chen2019gan, hayes2019logan, hilprecht2019monte}. An attack against semantic segmentation models was proposed by He et al. \cite{he2019segmentations}, where a shadow semantic segmentation model is trained, and is used to train a binary classifier. The classifier is trained on image patches, and final decision regarding the query image is set by aggregation of the per-patch classification scores. The input to the classifier is a structured loss map between the shadow model's output and the ground truth segmentation map. Although this task is the closest to ours, our work is the first study of membership inference attacks on image translation models. We also note that \cite{he2019segmentations} consider the setting where other input-output samples from the data distribution (or a very similar distribution) are available, whereas our attack does not require this information.

Besides membership inference attacks, other privacy attacks against neural networks exist. We refer the reader to the supplementary material (SM) for more details.
\subsection{Conditional Image Generation}

Image-to-image translation is the task of mapping an image from a source domain to a target domain, while preserving the semantic and geometric content of the input image. Currently, the most popular methods for training image-to-image translation models use Generative Adversarial Networks (GANs) \cite{goodfellow2014generative} and are currently used in two main scenarios:
(i) unsupervised image translation between domains \cite{zhu2017unpaired, kim2017learning, liu2017unsupervised, choi2018stargan};
(ii) serving as a perceptual image loss function \cite{isola2017image, wang2018high, zhu2017toward}. 
In this work we introduce the novel task of MIA on conditional image generation models.

\subsection{Semantic Segmentation}
Semantic segmentation is the task assigning a class label to each pixel in the input image. This can be thought of a classification problem for each pixel. State-of-the-art methods \cite{xiao2018unified, sun2019high} are based on fully convolutional networks and multi-scale representations of the input \cite{liu2019recent, minaee2020image}.

\section{Difficulty-based MIA}

We investigate the effect of task difficulty and dimensionality on the success of MIA.
Consequently, we concentrate on two promising tasks for MIA, image translation and semantic segmentation. We also present a novel image predictability error which significantly improves MIA accuracy.

\subsection{Effects of task difficulty and dimensionality}
\label{sec:invesigation_of_tasks}

Every neural network model is a potential target for MIA. Previous work attempted MIA on many different models (classification, GANs, segmentation) with highly variable accuracy. In this section we present an investigation into two factors that affect MIA accuracy: task difficulty and output dimensionality. Full details are provided in Sec.~\ref{sec:investigation_experimetns}. We perform reconstruction-based MIA by measuring the reconstruction error between the model output and the ground truth. This is done for multiple models, datasets and tasks. The attack method is described in more detail in Sec.~\ref{sec:main_method}.

\begin{table}[tb]
\centering
\small
    \begin{tabular}{lcc}
    \toprule
    \textbf{Model} & \textbf{Task} & \textbf{Reconstruction Error}\\
    \midrule
    NVAE & CelebA2Self  &  50.74\% \\
    \midrule
    Pix2pixHD & Maps2sat & 95.74\% \\
    Pix2pixHD & Cityscapes & 96.04\%\\
    \midrule
    Pix2pixHD & Landmarks2CelebA & 99.54\% \\

	\bottomrule
    \end{tabular}

\vspace{0.25cm}
\caption{Comparison of reconstruction-based MIA accuracy on tasks with different difficulties. Easier tasks, e.g.\ CelebA2Self, in which there no uncertainty in the output given the input image, suffer less from memorization of the training data and therefore have lower vulnerability to MIA. As the uncertainty increases (segmentation maps and landmarks) models tend to memorize the training data and therefore the MIA accuracy increases. }

\label{tab:mia_diff_tasks}
\end{table}

\textbf{MIA accuracy vs. task difficulty:} We present results of reconstruction-based attack on three different tasks of different difficulties. We define task difficulty as the uncertainty in the output given the input image. The tasks are: i) auto-encoding - translating an image to itself. ii) Segmentation-to-image translation. iii) Landmark to face translation. The first task is the easiest as the output is trivially determined by the input. Landmark-to-face is harder than segmentation-to-image as the input contains less information on the output (e.g. no information on the identity of the face requires much more memorization). The results are presented in Tab.~\ref{tab:mia_diff_tasks}, where it can be seen that indeed MIA performance is more accurate for harder tasks.

\begin{figure}[t]
\begin{center}

\subfigure[Pix2pixHD]{
\includegraphics[width=0.46\linewidth, height=0.4\linewidth]{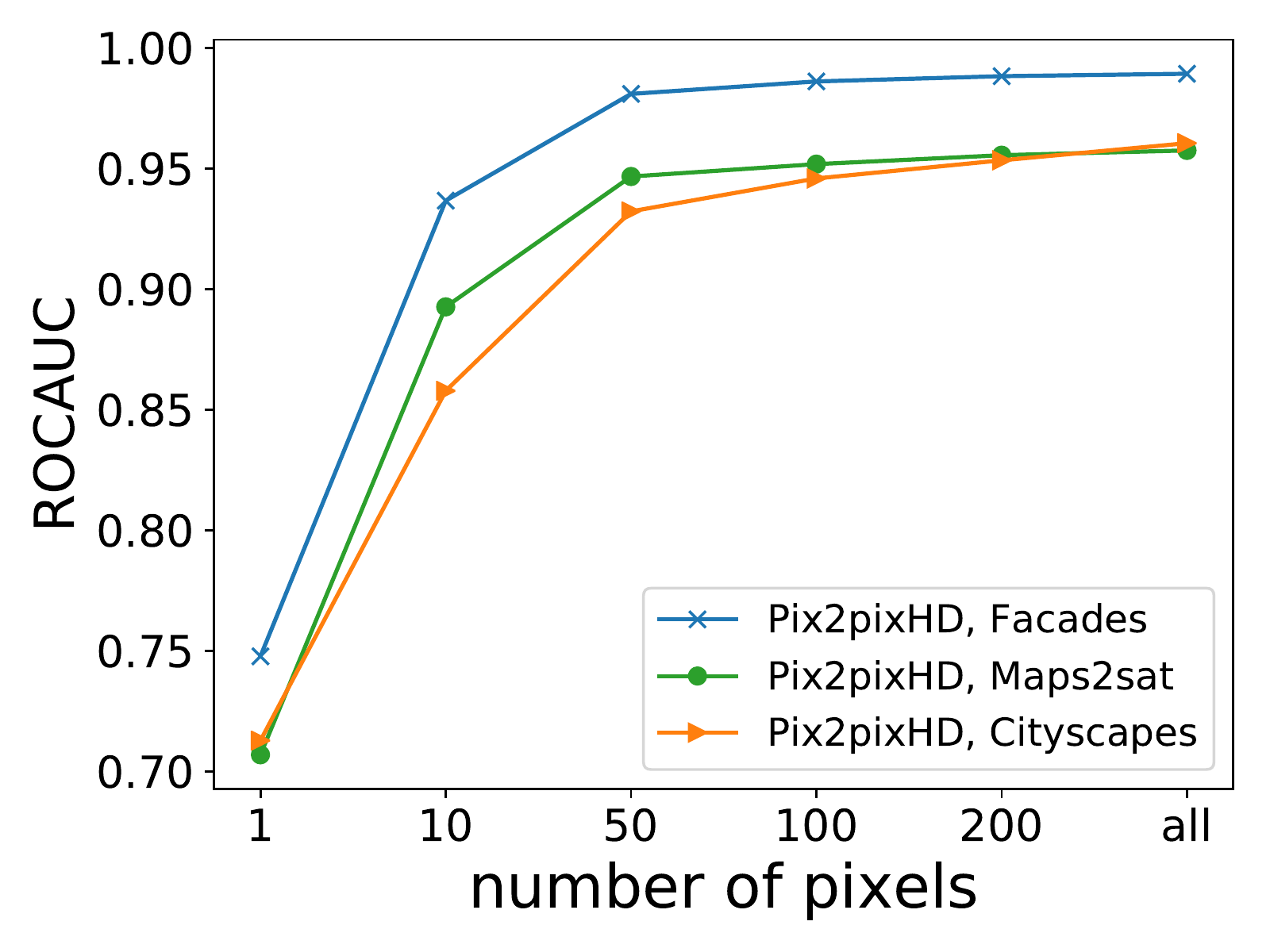}
}
\subfigure[Semantic Segmentation]{
\includegraphics[width=0.46\linewidth, height=0.4\linewidth]{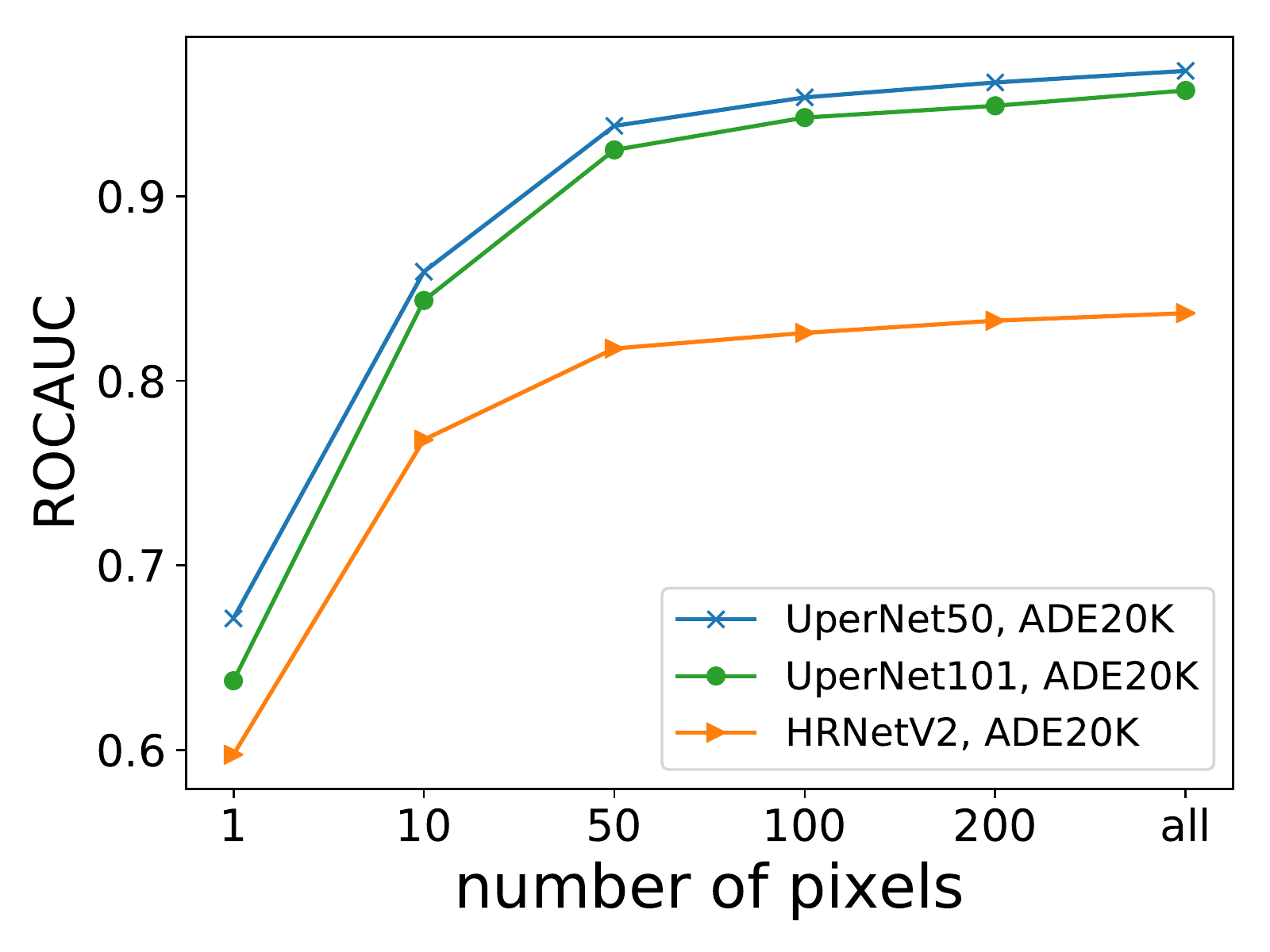}
}

\end{center}
\caption{Effect of reducing output dimensionality over a reconstruction-based attack. MIA accuracy is correlated with the output dimension, i.e.\ number of pixels, demonstrating that high output dimensionality tasks are more vulnerable to MIA.}
\label{fig:dim_reduce_effect}
\end{figure}

\textbf{MIA accuracy vs. output dimensionality:} Many MIA approaches attack classification networks that have only a single output, usually a probability vector or in the more restrictive case, a single label. It is natural to ask if tasks with higher dimensional outputs are more vulnerable to MIA due to the ensemble effect of attacking each individual output dimension. The intuition behind this is similar to the ensemble effect of boosting algorithms - each output pixel serves as a weak attack, and the aggregation of the reconstruction errors between all output pixels can produce a strong attack. In Fig.~\ref{fig:dim_reduce_effect}, we provide a comparison of reconstruction-based MIA when subsets of different sizes are used as the output. Note that segmentation with only a single pixel output is equivalent to classification. We can observe that MIA accuracy indeed scales with output dimensionality.

\begin{figure*}[t]
\begin{center}
\begin{tabular}{@{\hskip0pt}c@{\hskip0pt}c@{\hskip3pt}c@{\hskip0pt}c@{\hskip2pt}c@{\hskip0pt}c@{\hskip2pt}c@{\hskip0pt}c}
& & City - input & City - out & Maps - input & Maps - out & Covid - input & Covid - out \\
\begin{turn}{90} ~~~~~~~~ Low  \end{turn} &
\begin{turn}{90} ~ predictability  \end{turn} &
\includegraphics[width=0.14\linewidth, height=0.14\linewidth]{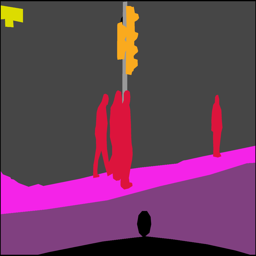} &
\includegraphics[width=0.14\linewidth, height=0.14\linewidth]{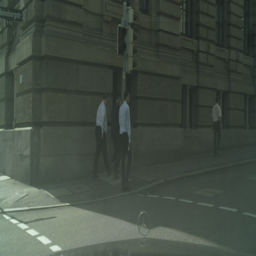} &
\includegraphics[width=0.14\linewidth, height=0.14\linewidth]{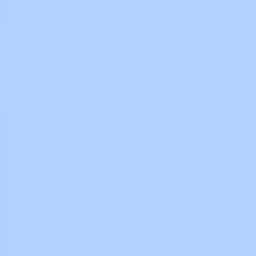} &
\includegraphics[width=0.14\linewidth, height=0.14\linewidth]{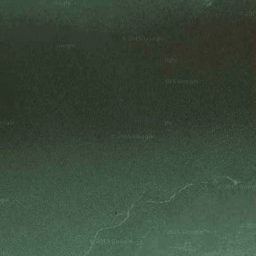} &
\includegraphics[width=0.14\linewidth, height=0.14\linewidth]{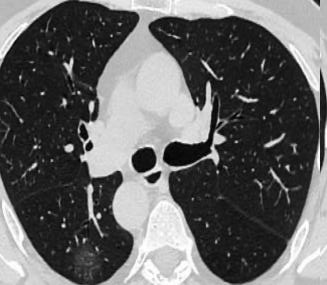} &
\includegraphics[width=0.14\linewidth, height=0.14\linewidth]{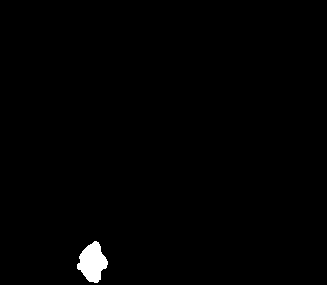} \\
\begin{turn}{90} ~~~~~~~~ High \end{turn} &
\begin{turn}{90} ~ predictability  \end{turn} &
\includegraphics[width=0.14\linewidth, height=0.14\linewidth]{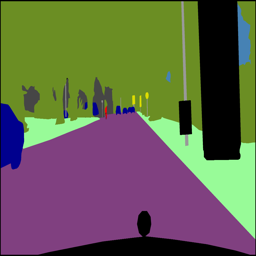} &
\includegraphics[width=0.14\linewidth, height=0.14\linewidth]{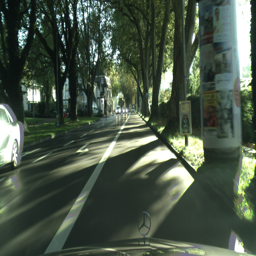} &
\includegraphics[width=0.14\linewidth, height=0.14\linewidth]{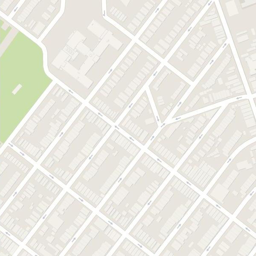} &
\includegraphics[width=0.14\linewidth, height=0.14\linewidth]{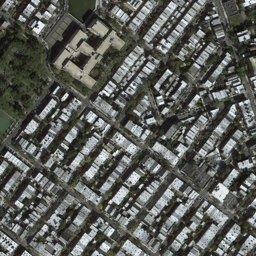} &
\includegraphics[width=0.14\linewidth, height=0.14\linewidth]{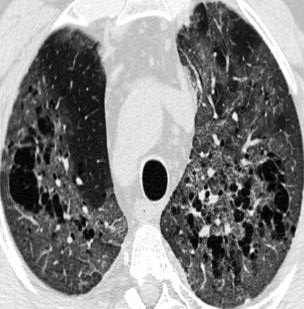} &
\includegraphics[width=0.14\linewidth, height=0.14\linewidth]{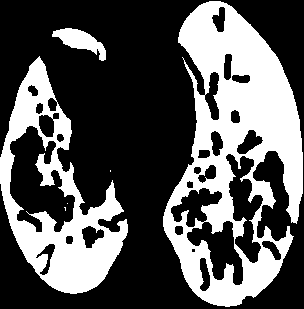} \\
\end{tabular}
\end{center}
\caption{Examples of input-output pairs from the Cityscapes, Maps2sat and COVID19-CT datasets that received the lowest (first row) and highest (second row) predictability errors using our single-sample approach. It can be seen that detailed images with complicated patterns are ranked as difficult to predict, while images with less details and lower contrast are ranked as easier to predict.}
\label{fig:diff_score_samples}
\end{figure*}

\subsection{MIA for high output dimensionality}
\label{sec:main_method}

We showed in Sec.~\ref{sec:invesigation_of_tasks} that both task difficulty and output dimensionality are correlated with the accuracy of MIA. We therefore focus on two important but difficult image tasks that have high-dimensional outputs: image translation and semantic segmentation. To the best of our knowledge, this is the first paper to consider MIA on image translation models.

We propose a simple but effective attack, assuming that the attacker has only a black-box access to the victim model $\mathbf{V}$.
Differently from most previous works, we do not use shadow models or train a binary classifier, and thus do not require any additional training data and query the victim model only once. 

Our membership attack is performed on a pair of query images $(x, y)$ where $x \in \mathbb{R}^{h \times w}$ is an image from the input domain ($h$ and $w$ are the image height and width respectively)  and $y \in \mathbb{R}^{h \times w}$ is the ground truth from the output domain. The requirement of the availability of the ground truth image $y$ is in-line with previous works, and is a reasonable assumption in our target scenario. For each query we compute a membership error, $L_{mem}$ (see Eq.~(\ref{eq:L-mem})), to which we apply a pre-defined threshold $\tau$, such that all queries where $L_{mem}(x, y) < \tau$ are marked as members of the training data. The membership error has two elements: reconstruction error and predictability error.

\vspace{-6pt}
\subsubsection{Reconstruction Error for Membership}
\label{sec:l_rec_method}

Typical MIA on classification models consider the probability (or confidence) given by the model to the correct class. Semantic segmentation is an extension where the output is a probability vector for each pixel. Image translation models are different as they output a color value of each pixel. This value is the maximum likelihood estimate, and no probability distribution over possible values is given. 

We propose to use the loss term as a reconstruction error, $L_{rec}$, to compute the pixel-wise difference between the output produced by a black-box access to the model, $\mathbf{V}(x)$, and the ground truth $y$. For semantic segmentation, where the output is a probability vector for each pixel, we use the cross-entropy error. For medical segmentation, we use the weighted IoU (Intersection over Union) loss and binary cross entropy loss. In the case of image-translation we use the $L_1$ error as we do not assume access to the discriminator and therefore can not use a GAN loss. 

Due to memorization during the training process, the model output typically has lower reconstruction errors for images in the training set compared to unknown images.

\subsubsection{Predictable and Unpredictable Images}
\label{sec:diff_score}

In this section we address the following question: Given an input-output sample, is the output easily predictable from the input. Consider, for example, supervised segmentation-to-image translation. I.e., the task is to "invert" the segmentation process, and recover the original image that gave rise to a given segmentation map. It is clear that not all cases are equally predictable: (i) hard to predict images have sharp and detailed textures, whereas more predictable images have blurrier textures; (ii) images with semantic segmentation maps that contain only few categories provide less guidance than those with more detailed segmentation maps, making the correct prediction less certain. The image predictability error should quantify these observations. In Sec.~\ref{sec:main_method_results} we show that such a predictability error is important for increasing accuracy of membership inference attacks.

We briefly describe two previous approaches for measuring image difficulty:

\textbf{Human-Supervised:} Ionescu et al. \cite{tudor2016hard} proposed to define image difficulty as the human response time for solving a visual search task. For this, they collected human annotations for the PASCAL VOC 2012 dataset \cite{Everingham10} and trained a regression model, based on pre-trained deep features, to predict the collected difficulty score. The disadvantage of this method is that human-specified difficulty scores may not correlate to the predictability of the image synthesis by neural networks. This is demonstrated empirically in Sec.~\ref{sec:supervised_results}.

\textbf{Multi-Image:} Another approach taken by \cite{chen2019gan} is training a model on a set of image pairs similar to the target distribution. This approach uses the reconstruction error of the external model on the target image pairs as its predictability error - larger reconstruction errors correspond to harder to predict images. This approach has a significant drawback: a large number of images, similar to the target image, are required in order to learn a reliable generative model. In many cases, images from the target distribution may not be available. Additionally, training a model for every task is tedious and computationally expensive. 

\textbf{The Proposed Single-sample predictability error:} We propose a new method to assign a predictability error for models with image outputs. This error measures the accuracy of predicting the output pixel from its high-level representation using linear regression. 

A related approach was proposed by Hacohen et al. \cite{hacohen2019power} for measuring image difficulty for classification models. Our method is significantly different as it is trained on a single input-output sample rather than on a large dataset.

The linear regression model uses image features of a pre-trained Wide-ResNet$50{\times}2$ \cite{zagoruyko2016wide}. We concatenate the activation values in the first $4$ blocks, giving $56{\times}56$ feature vectors of size $3840$ each. The output image resolution is reduced to $56{\times}56$ to match the size of the first Wide-ResNet$50{\times}2$ block.
We denote the concatenated feature vector for pixel $i$ as $\psi(i)$. 

The linear regression model $\mathbf{P}$ is a matrix of size $3840{\times}3$, multiplied with the feature vector $\psi(i)$ of pixel $i$ to give an estimate of the RGB colors $y^i$. We optimize $\mathbf{P}$ over 70\% randomly selected pixels. The image predictability error is the average absolute error over the 30\% unselected pixels:
\begin{equation}
\label{eq:L-diff}
L_{pred}(x, y) = \frac{1}{N}\sum_{i=1}^{N} \| \mathbf{P} \psi (i) - y^i\| _1
\end{equation}
where $y^i$ is the ground truth value of the $i_{th}$ pixel in the resized ground truth image $y$. Fig.~\ref{fig:diff_score_samples} presents some images that received the highest and lowest predictability errors. See supplementary material (SM) for more details.

\subsubsection{Membership Error}

As observed before, for some samples the output images can be easily predicted from the input, while for other samples the output can not be predicted. While the reconstruction error achieves high MIA success rates, it has a significant limitation - it does not discriminate between predictable and unpredictable samples. The victim model will have higher errors when generating unpredictable training samples, and lower errors when generating easily predictable ones. In such samples reconstruction error may result in wrong membership classification.

Given our image predictability error $L_{pred}$ in Eq.~(\ref{eq:L-diff}) and the reconstruction error $L_{rec}$ we calculate a membership error $L_{mem}$ as follows:
\begin{equation}
\label{eq:L-mem}
L_{mem}(x,y) = L_{rec}(x, y) - \alpha \cdot L_{pred}(x, y)
\end{equation}
$L_{mem}$ is computed by subtracting the predictability error $L_{pred}$ from the reconstruction error $L_{rec}$ weighted by $\alpha$. Unless specified otherwise, we use $\alpha=1.0$, and present the effect of different $\alpha$ values in the SM. This lowers the membership error $L_{mem}$ for harder-to-predict images compared to easier-to-predict images having the same reconstruction error. See Fig.~\ref{fig:method_plot} for an overview illustration of our method.

Using the membership error $L_{mem}$ (\ref{eq:L-mem}) for MIA substantially improves the success rates in all of our experiments, as shown in Tab.~\ref{tab:main_method_results} and Fig.~\ref{fig:calibration_effect}

\section{Experiments}
We first investigate two factors that affect MIA accuracy, task difficulty and output dimensionality, and show that MIA attacks are easier on difficult tasks with high output dimensionality. We then extensively evaluate our MIA on image translation and semantic segmentation networks. We also compare our single-sample predictability error against strong baselines. Additional results and ablations can be found in the SM. In accordance with previous membership attack works, the success rate is measured using the area under the ROC curve (ROCAUC) metric.

\subsection{Effectiveness of membership inference attacks}
\label{sec:investigation_experimetns}
As mentioned in Sec.~\ref{sec:invesigation_of_tasks}, there has been extensive research on MIA against various neural networks, resulting in variable accuracy. We investigated two factors that affect MIA accuracy: task difficulty and output dimensionality. 

\subsubsection{MIA accuracy vs. task difficulty}

We defined task difficulty as the uncertainty of the output given the input image. In the limit of sufficiently large training datasets, when models are trained to perform easy tasks, such as auto-encoding - translating an image to itself, they are able to generalize well to unseen images, and do not need to depend on memorization of the training data in order to minimize the loss function. As membership inference attacks are highly correlated with model memorization, their performance decreases on such tasks. Similarly, models struggle with learning difficult tasks, in which \ul{the input does not contain sufficient information to fully specify the output}, and therefore the loss minimization encourages the model to memorize the training samples. This lack of predictability acts as a strong motivation for memorization, even at the limit of well-trained models trained on large datasets, provided sufficient capacity.

In order to demonstrate this, we performed reconstruction-based MIA, by using $L_{rec}$ described in Sec.~\ref{sec:l_rec_method}, on three tasks of different levels of difficulty. The first and easiest task is auto-encoding. For this, we attack a NVAE \cite{vahdat2020nvae} model, trained on the CelebA \cite{liu2015faceattributes} dataset. The second task, more difficult than auto-encoding, is segmentation-to-image translation. We attack two Pix2PixHD models \cite{wang2018high}, trained on the Maps2sat \cite{isola2017image} and Cityscapes \cite{Cordts2016Cityscapes} datasets. The third and most difficult task is landmarks-to-face translation. For this task we extracted facial landmarks \cite{bulat2017far} from $50$K CelebA images \cite{liu2015faceattributes}. We consider this to be the most difficult task out of the three as the input contains less information on the output in comparison to segmentation maps (e.g.\ no information regarding the identity of the face).
Results, presented in Tab.~\ref{tab:mia_diff_tasks}, demonstrate that reconstruction-based MIA are more successful on difficult tasks.

\begin{figure}[tb]
\begin{center}
\subfigure[Pix2pixHD - Maps]{
\includegraphics[width=0.47\linewidth, height=0.35\linewidth]{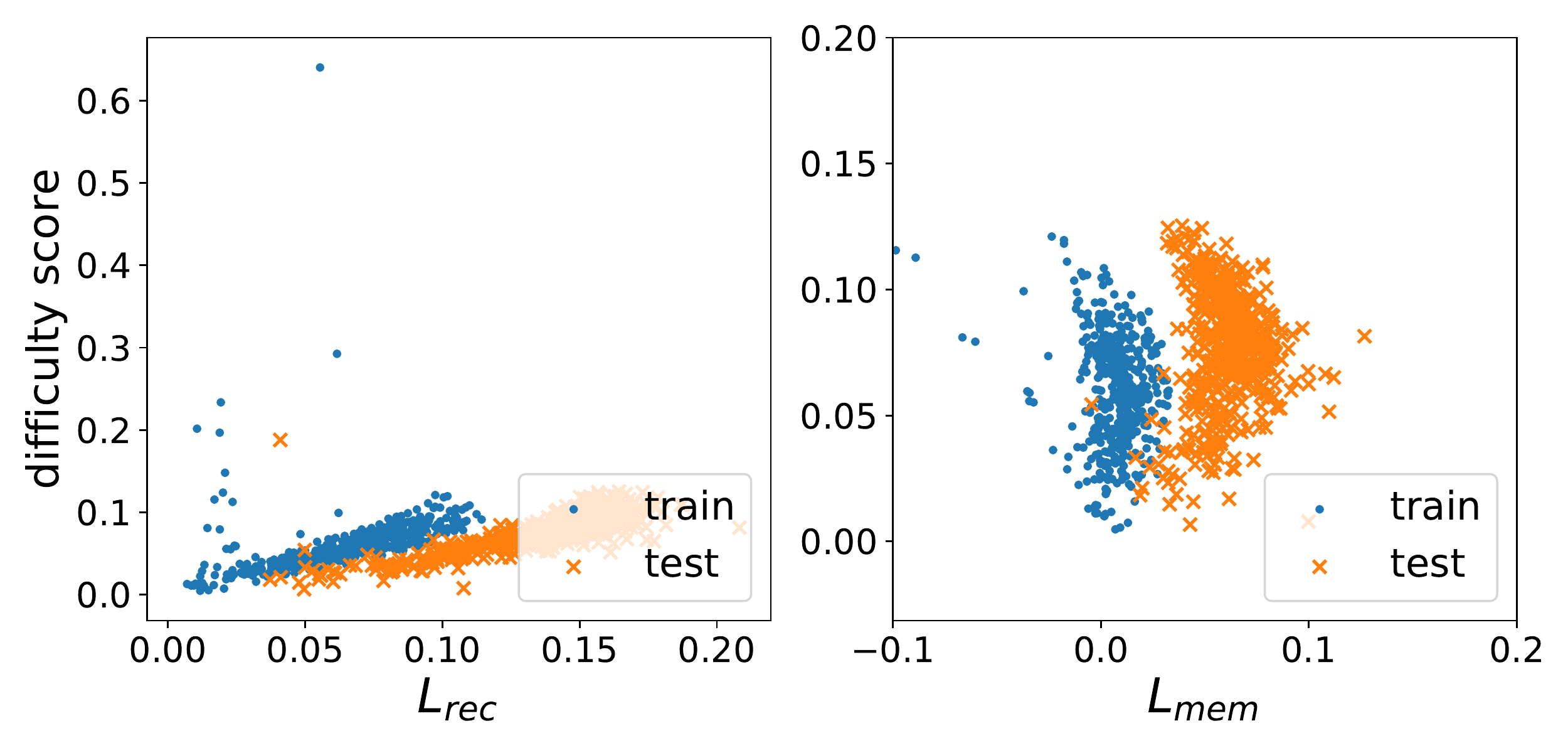}
}
\subfigure[Pix2pixHD - Cityscapes]{
\includegraphics[width=0.47\linewidth, height=0.35\linewidth]{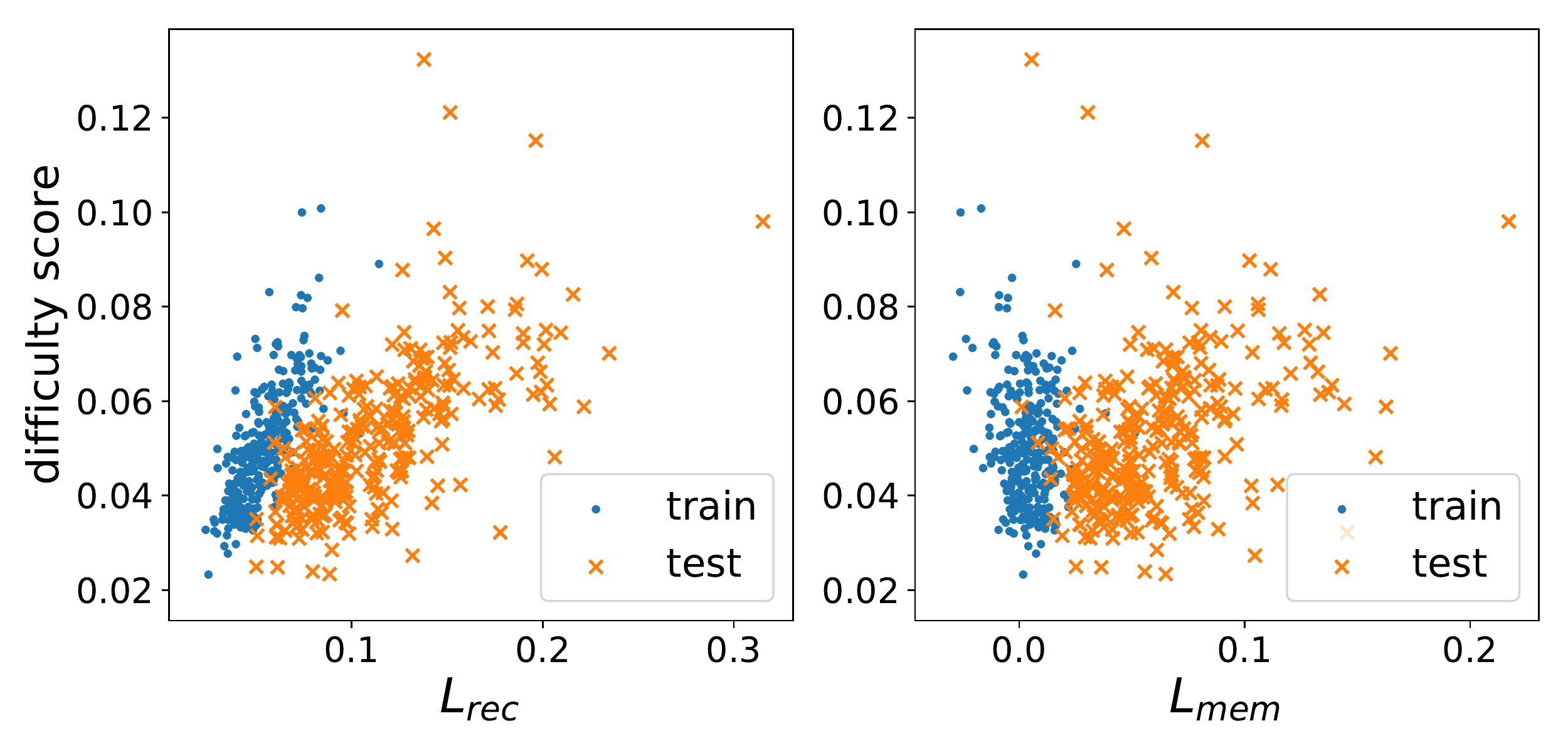}
}
\end{center}
\caption{The proposed membership error $L_{mem}$ can better separate train (blue) and test (orange) images by a simple threshold (i.e. a vertical line) compared to the reconstruction error $L_{rec}$.}
\label{fig:calibration_effect}
\end{figure}

\vspace{-6pt}
\subsubsection{MIA accuracy vs. output dimensionality}

Previous works mostly focused on MIA against classification models, where there is a single output, i.e.\ probability vector or in the more restrictive case, a single label. It is natural to ask whether higher dimensional outputs are more vulnerable due to the ensemble effect of combining the attacks on each individual output dimension to a stronger, joint attack.

We perform reconstruction-based MIA on the Pix2PixHD architecture, trained on the CMP Facades \cite{Tylecek13}, Maps2Sat and Cityscapes datasets, as well as on three semantic segmentaion models - UperNet50, UperNet101 \cite{xiao2018unified} (using ResNet50 and ResNet101 as backbones) and HRNetV2 \cite{sun2019high} - trained on the ADE20K dataset \cite{zhou2017scene}. Fig.~\ref{fig:dim_reduce_effect} demonstrates the effect of reducing the output dimension on the attack accuracy. The reduction is achieved by randomly sampling $N$ output pixels, and using them as the output, where $N$ ranges from a single pixel and up to $200$ pixels. Note that in the case of semantic segmentation, having only a single pixel output is equivalent to classification. We repeat this experiment $10$ times and report the average result. 

We observed that MIA accuracy indeed scales with the number of output dimensions. Results for other models are presented in the SM. 

\subsection{MIA accuracy evaluation}
\label{sec:main_method_results}
We evaluate our membership attack on three image translation architectures, Pix2Pix \cite{isola2017image}, Pix2PixHD \cite{wang2018high}, SPADE \cite{park2019semantic}, three semantic segmentation architectures - UperNet50 and UperNet101 \cite{xiao2018unified} (ResNet50 and ResNet101 as backbones), HRNetV2 \cite{sun2019high} as well as two medical segmentation architectures - Inf-Net \cite{fan2020inf} and PraNet \cite{fan2020pra}.
We evaluated on various datasets, including CMP Facades \cite{Tylecek13}, Maps2sat \cite{isola2017image}, Cityscapes \cite{Cordts2016Cityscapes}, ADE20K \cite{zhou2017scene}. 

In the case of medical segmentation we evaluated two tasks: lung infection segmentation from Covid-19 CT images \cite{fan2020inf} and polyp segmentation in colonoscopy images \cite{silva2014toward, bernal2015wm, tajbakhsh2015automated, vazquez2017benchmark, jha2020kvasir}. 

 All pix2pix and pix2pixHD models are trained from scratch, with the exception of the Cityscapes dataset on the Pix2pixHD architecture in which we use the supplied large pre-trained model for computational constraints on the high resolution. The rest of the models are pre-trained.

It can be seen in Tab.~\ref{tab:main_method_results} that while using the reconstruction error alone achieves a high success rate, the membership error (which calibrates the result by sample predictability) significantly improves the results. 
Fig~\ref{fig:calibration_effect} demonstrates the effect of subtracting the predictability error from the reconstruction error. After calibration, a single threshold on the membership error can separate train and test images. Results on additional benchmarks are presented in the SM.

Utilizing common image augmentations, i.e.\ horizontal flipping and random cropping, in order to construct a larger a larger set $\{(x_{aug}, y_{aug})\}$ has a small impact over the attack accuracy, as the output dimension is large enough as is.


\begin{table}[tb]
\centering
\small
    \begin{tabular}{lccccc}
    \toprule
    \textbf{Model} & \textbf{Dataset} & \multicolumn{2}{c}{\textbf{Single}} & \multicolumn{2}{c}{\textbf{Multi}}\\
     &  & \textbf{Ours} & \textbf{Superv.} &  \textbf{In-Dist.}\\
    \midrule
    Pix2pix & Facades & \textbf{97.59\%} & 93.67\%  & -\\
    Pix2pix & Maps2sat & 85.65\% & 86.48\%  & \textbf{92.43\%}\\
    Pix2pix & Cityscapes & \textbf{83.23\%} & 77.06\%  & 82.47\%\\
    \midrule
    Pix2pixHD & Facades & \textbf{99.95\%} & 98.86\% & -\\
    Pix2pixHD & Maps2sat & \textbf{99.42\%} & 98.38\%  & 82.87\%\\
    Pix2pixHD & Cityscapes & \textbf{99.09}\% & 96.86\%  & 94.76\%\\
    \midrule
    UperNet50 & ADE20K & \textbf{98.09\%} & 96.79\% & 79.47\% \\
    UperNet101 & ADE20K & \textbf{96.94\%}  & 95.49\% & 76.01\% \\
    HRNetV2 & ADE20K & \textbf{85.92\%} &  84.42\% & 84.48\%\\
	\bottomrule
    \end{tabular}
\vspace{0.25cm}
\caption{MIA accuracy of our self-supervised single-sample method vs. using human-supervised single-sample and multi-image baselines for the predictability error. Note that in-distribution multi-image requires extra supervision of $100$ images}

\label{tab:single_vs_multi_diff_score}
\end{table}

\subsubsection{Comparison to Human Supervision}
\label{sec:supervised_results}
We compare our self-supervised single-sample predictability error with the human-supervised difficulty score described in Sec.~\ref{sec:diff_score}. This score was proposed by Ionescu et al. \cite{tudor2016hard}, which defined image difficulty to be the human response time for solving a visual search task. In order to provide a fair comparison, we replace the pretrained VGG-f \cite{chatfield2014return} features, used by \cite{tudor2016hard}, with the more powerful pretrained Wide-ResNet$50\times2$ \cite{zagoruyko2016wide} features we use in our predictability error. Samples of images ranked as easy and hard by the supervised score are presented in the SM. As can be seen in Tab.~\ref{tab:single_vs_multi_diff_score}, our self-supervised single-sample predictability error outperforms the human-supervised difficulty score. 
In the SM, we provide a comparison of the relation between the reconstruction error and the supervised score to the relation between the reconstruction error and our self-supervised predictability error, and show that our score is better correlated to the reconstruction error.

\begin{table}[tb]
\centering
\small
    \begin{tabular}{lccc}
    \toprule
    \textbf{Model} & \textbf{Dataset} & \textbf{Ours} & \textbf{Shadow Model}\\
    \midrule
    Pix2pix & Maps2sat & \textbf{85.65\%} & 80.15\%\\
    Pix2pix & Cityscapes & \textbf{83.23\%} & 78.68\%\\

    \midrule
    Pix2pixHD & Maps2sat & \textbf{99.42\%} & 98.63\%\\
    Pix2pixHD & Cityscapes & \textbf{99.09\%} & 96.39\%\\
	\bottomrule
    \end{tabular}
\vspace{0.25cm}
\caption{Comparison between our MIA and the popular shadow-model-based classifier attack, using $100$ train and $100$ test samples. Our MIA outperforms while not requiring extra training images.}

\label{tab:shadow_model}
\end{table}

\subsubsection{Comparison to Multi-Image Scores}
\label{sec:single_vs_multi_diff_score}

Although our MIA method does not require the availability of multiple auxiliary samples from the target distribution or from a similar distribution, it is interesting to compare our single sample predictability error to methods that use multiple samples. We compute multi-sample predictability errors (MSPS) by training a "shadow" model to map the input to output images in the auxiliary samples. As an upper-bound on MSPS performance, the shadow model is given exactly the same architecture as used by the victim model (although this knowledge may not be available in practice). The MSPS is defined by the reconstruction error of the shadow model on the target sample. Two scenario were evaluated:

\textbf{In-distribution data:} In this setting the shadow model's data shares the distribution of the victim's training data, by being trained on $100$ randomly sampled image pairs from the test set of the corresponding dataset. Facades was not used as it did not have enough images. The results are presented in Tab.~\ref{tab:single_vs_multi_diff_score}. For Pix2PixHD and semantic segmentation, MSPS underperformed our method (as $100$ samples are insufficient for training such large models). As Pix2Pix is a smaller network, MSPS was more successful there, obtaining competitive results with our method. Note that it still requires extra samples, often not available. We analyzed the number of samples required for MSPS to reach the accuracy of our method, in most tasks, even 100 were insufficient (see SM).

\textbf{Auxiliary dataset:} As suggested by He et al. \cite{he2019segmentations}, we also compare our method to the setting were many out-of-distribution but similar samples are available. We trained shadow models on $4K$ image pairs from the BDD dataset \cite{yu2018bdd100k} as MSPS for the Cityscapes dataset, as both datasets consist of street scene images and have compatible label spaces. We found that MSPS underpeformed our method by $10\% - 30\%$. (see SM for exact results). Note that it is rare to have similar datasets with nearly identical labels. Cityscapes was the only dataset from those evaluated in this paper for which such a similar dataset could be found. 

\textbf{Shadow-model classifiers:} Although deviating somewhat from predictability errors, for the interest of completeness, we report the ROCAUC results of the popular approach of shadow-model classifiers for image translation MIA, see Tab.~\ref{tab:shadow_model} (classification accuracy is lower, see SM). We use the approach of He et al. \cite{he2019segmentations} and train a classifier to distinguish between the "loss maps" of the train and test auxiliary samples of the shadow model. The classifier is then used to score images of the target dataset as train or test (see \cite{he2019segmentations} for the complete details).

We show that this approach underperforms our method in both in-distribution and auxiliary dataset settings (exact results presented in the SM). It is surprising that shadow models do not perform well on image translation MIA as they are very effective for image segmentation MIA (as shown in \cite{he2019segmentations}). We believe the difference in performance can be explained by the fact that segmentation maps have similar distributions between datasets with similar label spaces while natural images have very different distribution - making membership classifiers on the image2seg task generalize better than the seg2image task. We note again, that such techniques requires the availability of auxiliary in-distribution samples or very similar datasets which is often not possible. For example, medical segmentation datasets are often quite small due to the sensitivity of the data and high cost of obtaining ground truth labels. The Covid-19 CT dataset \cite{fan2020inf} is composed of $50$ train and $50$ test samples, too small to apply a shadow model based attack.




\section{Discussion}
\label{sec:discussion}

\textbf{Effect of Memorization:} Membership inference attacks are closely related to memorization in the victim model. In order to better understand this relation, we measure the success of our attack under different levels of memorization. We do so by evaluating our attack on checkpoints saved at different epochs during the training of our image-translation victim models. We observed that as the training process progresses, the victim model memorizes the training data which results in higher attack success rates (See SM).

\begin{table}[tb]
\centering
\small
    \begin{tabular}{lccc}
    \toprule
    \textbf{Model} & \textbf{Dataset} & \textbf{Orig} & \textbf{No $L_{rec}$}\\
    \midrule
    Pix2pix & Maps2sat & 84.22\% & 68.44\%\\
    Pix2pix & Cityscapes & 77.74\% & 51.06\% \\

    \midrule
    Pix2pixHD & Maps2sat & 95.74\% & 75.76\%\\
    Pix2pixHD & Cityscapes & 96.04\% & 56.58\%\\
	\bottomrule
    \end{tabular}
\vspace{0.25cm}
\caption{Effect of cGAN and reconstruction losses on the accuracy of reconstruction-based MIA. The cGAN loss is less susceptible to MIA.}

\label{tab:mia_rec_loss}
\end{table}





\textbf{Reconstruction loss effect on MIA:} We evaluated the effect of the reconstruction loss term on the accuracy of reconstruction-based MIA. For this, we compared the accuracy of the attack against image translation models, trained using both reconstruction and cGAN loss terms versus models that were only trained using the cGAN loss term. As can be seen in Tab.~\ref{tab:mia_rec_loss}, the reconstruction loss indeed has a significant impact over the attack accuracy. 

\begin{table}[tb]
\centering
\small
    \begin{tabular}{lccc}
    \toprule
    \textbf{Model} & \textbf{Dataset} & \textbf{No-defense} & \textbf{Argmax-defense}\\
    \midrule
    UperNet50 & ADE20K & 98.09\% & 98.05\%  \\
    UperNet101 & ADE20K & 96.94\%  & 97.11\%  \\
    HRNetV2 & ADE20K & 85.92\% &  89.32\% \\
	\bottomrule
    \end{tabular}
\vspace{0.25cm}
\caption{Comparison between our attack accuracy (ROCAUC) on undefended semantic segmentation models and models defended by the argmax defense. As can be seen, our attack still manages to succeed much better than random guessing.}

\label{tab:argmax_defense}
\end{table}

\textbf{Argmax defense:} In the argmax defense, the victim model returns only the predicted label, rather then the full probability vector. As image translation models predict pixel values, and does not output probability vectors, this defense does not apply. 
For semantic segmentation models, we evaluate our attack against this defense, by replacing the cross-entropy loss in $L_{rec}$ to the $L_0$ error. As can be seen in Tab.~\ref{tab:argmax_defense}, the attack efficacy is not reduced, demonstrating the weakness of this defense.  

\textbf{Differential private SGD (DP-SGD) defense}: In the defense by Abadi et al. \cite{abadi2016deep}, the commonly used Stochastic Gradient Descent optimization algorithm is modified in order to provide a differentially private \cite{dwork2014algorithmic} model. This is done by adding Gaussian noise to clipped gradients for each sample in every training batch. There exists a trade-off between privacy and utility, in which the amount of added noise must be large enough to ensure privacy while not degrading the model's outputs to the point where the model is useless. Training a deep model with DP-SGD is an unstable process. We experimented with multiple common configurations, i.e. added noise ratios and maximal gradient clipping threshold, and were not able to find a configuration that yields visually satisfying results. Hence, although the DP-SGD defense is theoretically protecting the victim model against membership inference attacks, in practice we find it to be impractical against our attack as it results with total corruption of the victim model. 

\begin{figure}[tb]
\begin{center}
\subfigure[Pix2pixHD]{
\includegraphics[width=0.45\linewidth, height=0.4\linewidth]{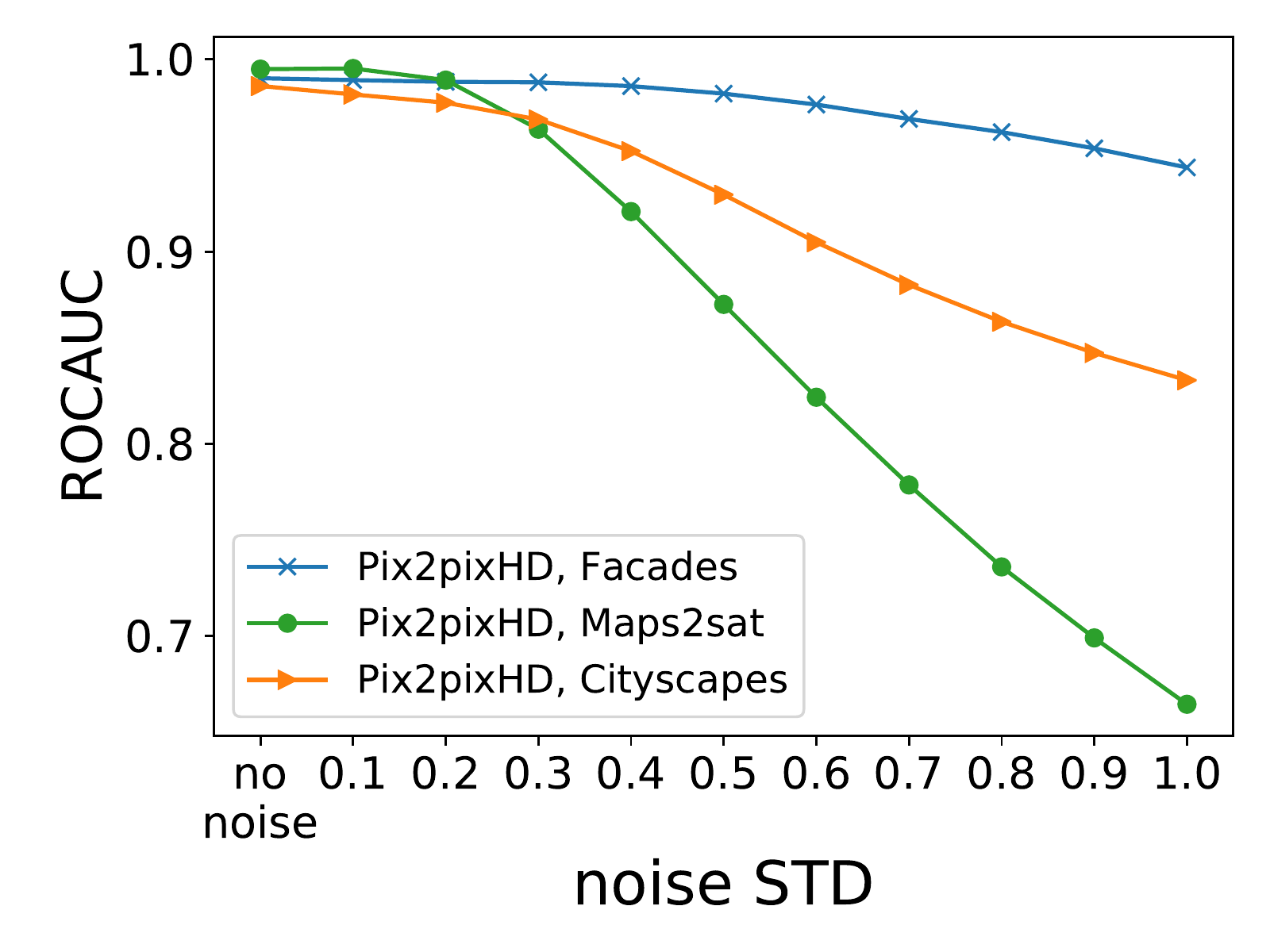}
}
\subfigure[Semantic Segmentaion]{
\includegraphics[width=0.45\linewidth, height=0.4\linewidth]{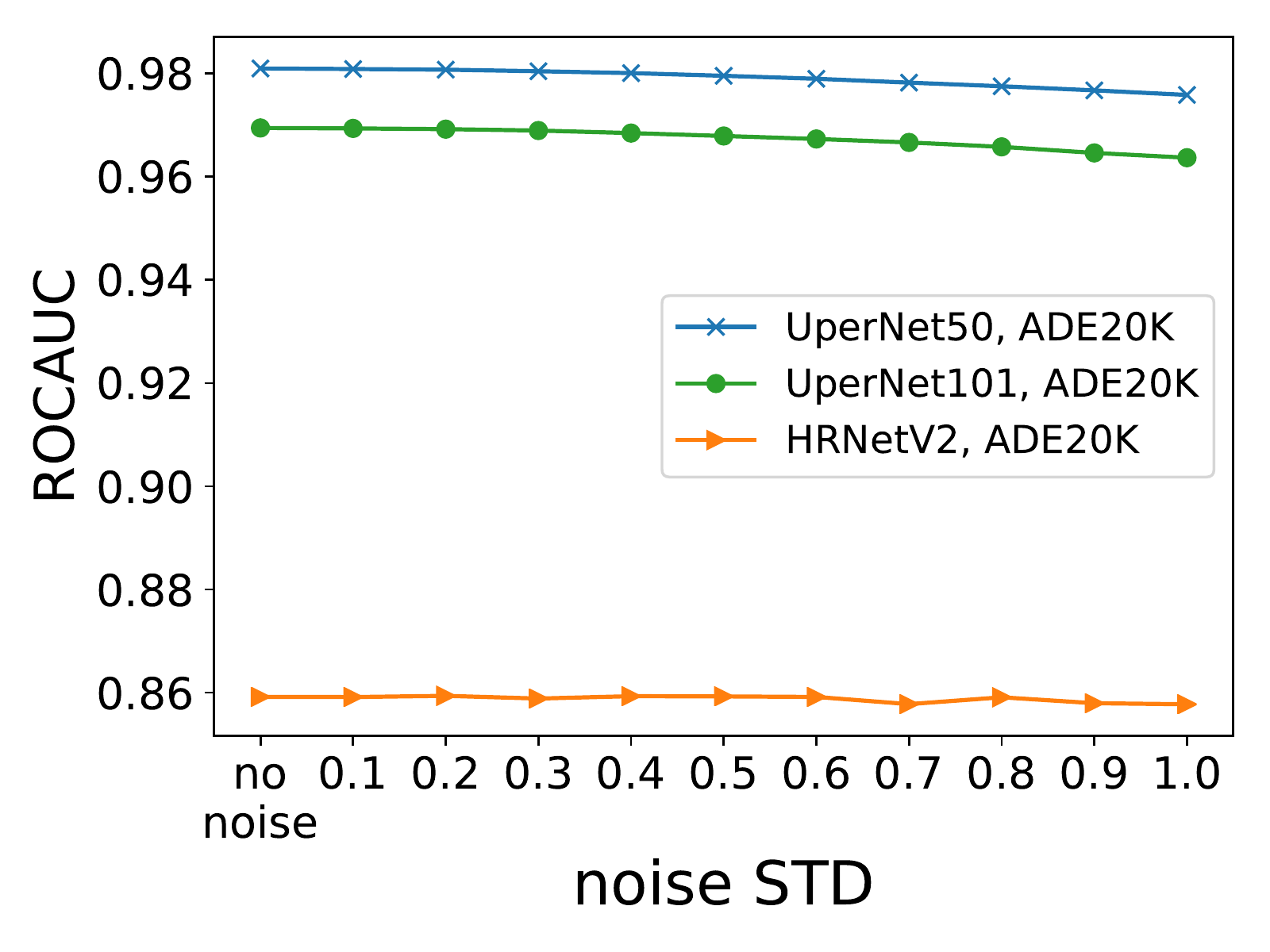}
}
\end{center}
\caption{Effect of Gauss defense on the attack accuracy. Even with large amounts of added noise, our attack still manages to succeed much better then random guessing.}
\label{fig:gauss_defense}
\end{figure}

\textbf{Gauss defense:} In this defense, we add Gaussian noise to the image generated by the victim model \cite{gilmer2019adversarial}. This attempts to hide specific artifacts of the model. We evaluate our attack accuracy as a function of different noise STD. Fig.~\ref{fig:gauss_defense} shows that a considerable amount of noise, which corrupts the generated output, is required in order to have a significant effect over our attack success. Moreover, it can be seen that even with large amounts of noise, our attack still manages to succeed much better than random guessing. This implies that our attack is robust to the Gauss defense. Results on additional benchmarks are presented in the SM.

\vspace{-6pt}
\section{Conclusion}

In this work, we highlight two properties that make tasks more vulnerable to MIA: i) Uncertainty: tasks where there is more uncertainty in the prediction of the output given an input ii) Output dimensionality: tasks with high-dimensional output. We show that a black-box reconstruction-based membership attack is very effective on two tasks that exhibit these properties: image translation and semantic segmentation, including medical segmentation. We further improve the attack by proposing a novel image predictability error. Our membership error, composed of the reconstruction and predictability errors, has been extensively evaluated on various benchmarks and was shown to achieve high accuracy.

\section{Acknowledgments}
This research was supported by grants from the Israel Science Foundation and from the DFG.

{\small
\bibliographystyle{ieee_fullname}
\bibliography{egbib}
}
\newpage
\clearpage
\section{Supplementary Material}
\subsection{Other Privacy Attacks}
\label{app:other_privacy_attacks}
Besides membership inference attacks, there exists a wide range of privacy attacks against neural networks. Model inversion attacks, first proposed by \cite{fredrikson2014privacy}, aim at reconstructing features of the training data, e.g.\ recovering an image of a person from face recognition models \cite{fredrikson2015model}. 
Property inference attacks, proposed by \cite{ganju2018property}, do not focus on the privacy of individual data samples, as in membership inference and model inversion attacks, but focus at inferring global properties of the training data, such as the environment in which the data was produced or the fraction
of the data that comes from a certain class. 

Model extraction attacks, also referred to as model stealing, attack a model $f$ by constructing a substitute model $\hat{f}$ that is either identical or equivalent to $f$ \cite{tramer2016stealing, jagielski2020high}. Related line of work \cite{wang2018stealing, oh2019towards} attempts to infer hyperparameters such as the optimization proccess, e.g.\ SGD or ADAM.

\subsection{Detailed Description of Our MIA Algorithm}
\label{app:algorithm}

Our MIA consist of computing the two terms in Eq.~(\ref{eq:L-mem}), i.e.\ $L_{rec}$ and $L_{pred}$ for a given query pair $(x,y)$, where $x$ is an image from the input domain and $y$ is the ground truth from the output domain, using only a black-box access to the victim conditional generation model $\mathbf{V}$.

$L_{rec}$ is computed using the pixel-wise error between the output image predicted by the model, $\mathbf{V}(x)$, and the ground truth image $y$, see step $1$ in the Algorithm 1. For image translation models, we set the pixel-wise error function, $err$, to be the $L_1$ loss:
\begin{equation}
\label{eq:L-rec_l1}
L^{trans}_{rec}(x, y) = \| \mathbf{V}(x) - y\|
\end{equation}
For semantic segmentation, where the output values are probability vectors rather then pixel values, we use the cross-entropy loss:
\begin{equation}
\label{eq:L-rec-seg}
L^{seg}_{rec}(x, y) = -log(\mathbf{V}(x)[y])
\end{equation}
In the case of medical segmentation, following Fan et al. ~\cite{fan2020pra, fan2020inf}, we use the weighted IoU loss and binary cross-entropy loss:

\begin{equation}
\label{eq:L-rec-med}
L^{med}_{rec}(x, y) = L^w_{IoU}(x,y) + L^w_{BCE}(x,y)
\end{equation}
 
Defined as: 
\begin{equation}
\label{eq:L-w-iou}
\small
L^w_{IoU} = 1 - \frac{\sum\limits_{i=1}^{H}\sum\limits_{j=1}^{W} w_{ij} (\mathbf{V}(x)_{ij}\cdot y_{ij})}{\sum\limits_{i=1}^{H}\sum\limits_{j=1}^{W} w_{ij} (\mathbf{V}(x)_{ij} + y_{ij} - \mathbf{V}(x)_{ij} \cdot y_{ij})}
\end{equation}

\begin{equation}
\label{eq:L-w-bce}
\small
L^w_{BCE} = - \frac{\sum\limits_{i=1}^{H}\sum\limits_{j=1}^{W} w_{ij} log(\mathbf{V}(x)[y]_{ij})}{\sum\limits_{i=1}^{H}\sum\limits_{j=1}^{W} w_{ij}}
\end{equation}

Where $H$ and $W$ are the height and width of the query sample, and $w_{ij}$ is the weight of pixel $(i, j)$ and is defined as follows, where $A_{ij}$ represents the area that surrounds the pixel $(i, j)$:
\begin{equation}
\label{eq:L-w-bce}
\small
w_{ij} = 1 - \left| \frac{\sum\limits_{m,n\in A_{ij}} y_{mn}}{\sum\limits_{m,n\in A_{ij}} 1} - y_{ij} \right|
\end{equation}

$L_{pred}$ is computed as the average error of a linear regression model, $\mathbf{P}$, in predicting pixel values from deep features of the input image. 

Our deep features are the activation values in the first $4$ blocks of a pre-trained Wide-ResNet$50{\times}2$ \cite{zagoruyko2016wide}. These features are of sizes $56{\times}56{\times}256$, $28{\times}28{\times}512$, $14{\times}14{\times}1024$, and $7{\times}7{\times}2048$.  We interpolate all features to size $56{\times}56$ using bi-linear interpolation (step 2), and also reduce the output image to  $56{\times}56$ using bicubic interpolation (step 3). This gives a concatenated feature vector of size $3840$ for each pixel $i$ in the $56{\times}56$ image ($256{+}512{+}1024{+}2048{=}3840$). We denote the concatenated feature vector for pixel $i$ as $\psi(i)$.

We randomly select $70\%$ of the pixels as train set, and compute a linear model $\mathbf{P}$ to estimate the RGB pixel values $y_{train}^i$ from the corresponding deep features $\psi_{train}(i)$ (step 4). The remaining $30\%$ of pixels will be used as a test split, $\{\psi_{test}, y_{test}\}$ (step 5). I.e.\ $|\psi_{train}| = 2195{\times}3840, |y_{train}| = 2195{\times}3$ and $|\psi_{test}| = 941{\times}3840, |y_{test}| = 941{\times}3$.

The linear regression model $\mathbf{P}$, a matrix of size $3840{\times}3$,
is trained to minimize the error over $\{\psi_{train}, y_{train}\}$ (step 6). $L_{pred}$ will be the average absolute error over $\{\psi_{test}, y_{test}\}$ (step 7). We found that fitting the linear model to 70\% of pixels and measuring the error on the remaining 30\% gives better results than just measuring the error of the linear fitting.

We compute $L_{mem}$ according to Eq.~(\ref{eq:L-mem}) and compare the results with a predefined threshold value $\tau$, such that any pair $(x, y)$ for which is holds that $L_{mem}(x, y) < \tau$ is denoted as a member of the victim models' $\mathbf{V}$ train set (steps 8-9). 

We have experimented with different resize methods (step 3) and found that our attack success rate is not very sensitive to the resize method. Additionally, we evaluated the effect of different train-test partitions (steps 4 \& 5) and found that using less than $50\%$ of the image pixels for training the linear regression model results with unstable performance, while all values of $50\%$ or above results in similar attack success rates.

\subsection{Parameter Selection}
\label{sec:alpha_effect}

\begin{figure}[h]
\begin{center}
\subfigure[Pix2pix]{
\includegraphics[width=0.46\linewidth, height=0.38\linewidth]{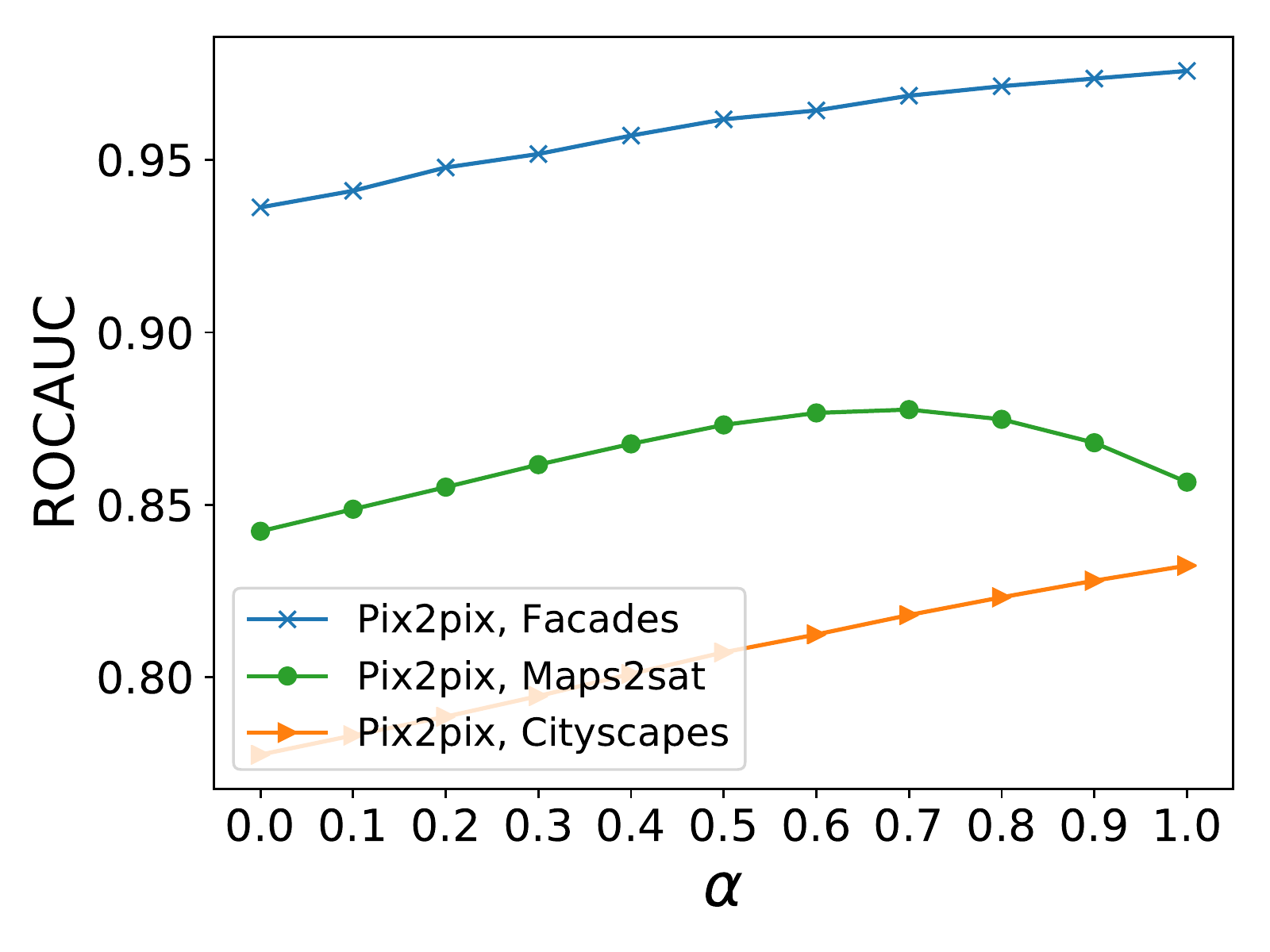}
}
\subfigure[Pix2pixHD]{
\includegraphics[width=0.46\linewidth, height=0.38\linewidth]{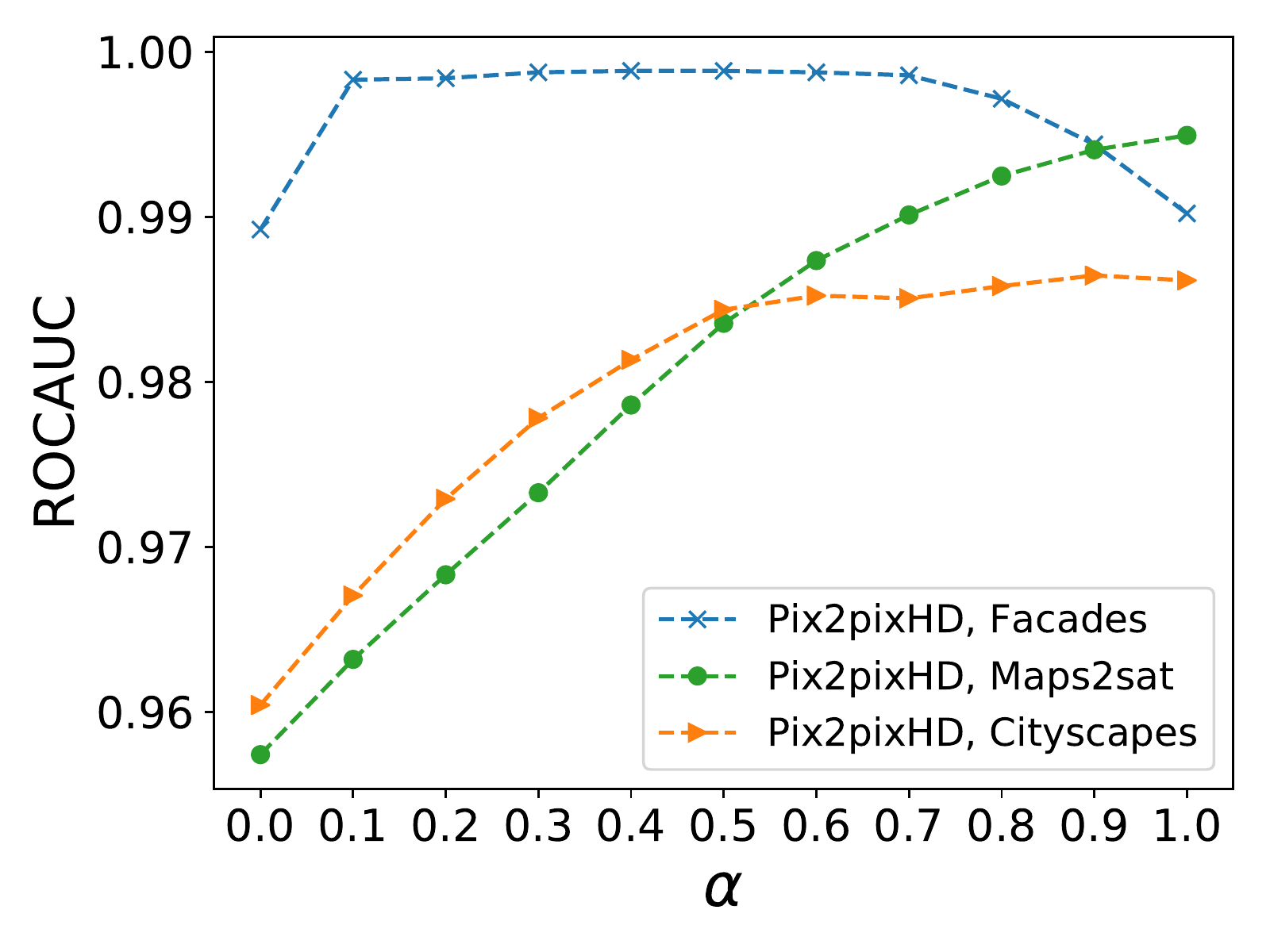}
}

\subfigure[Semantic Segmentation]{
\includegraphics[width=0.46\linewidth, height=0.38\linewidth]{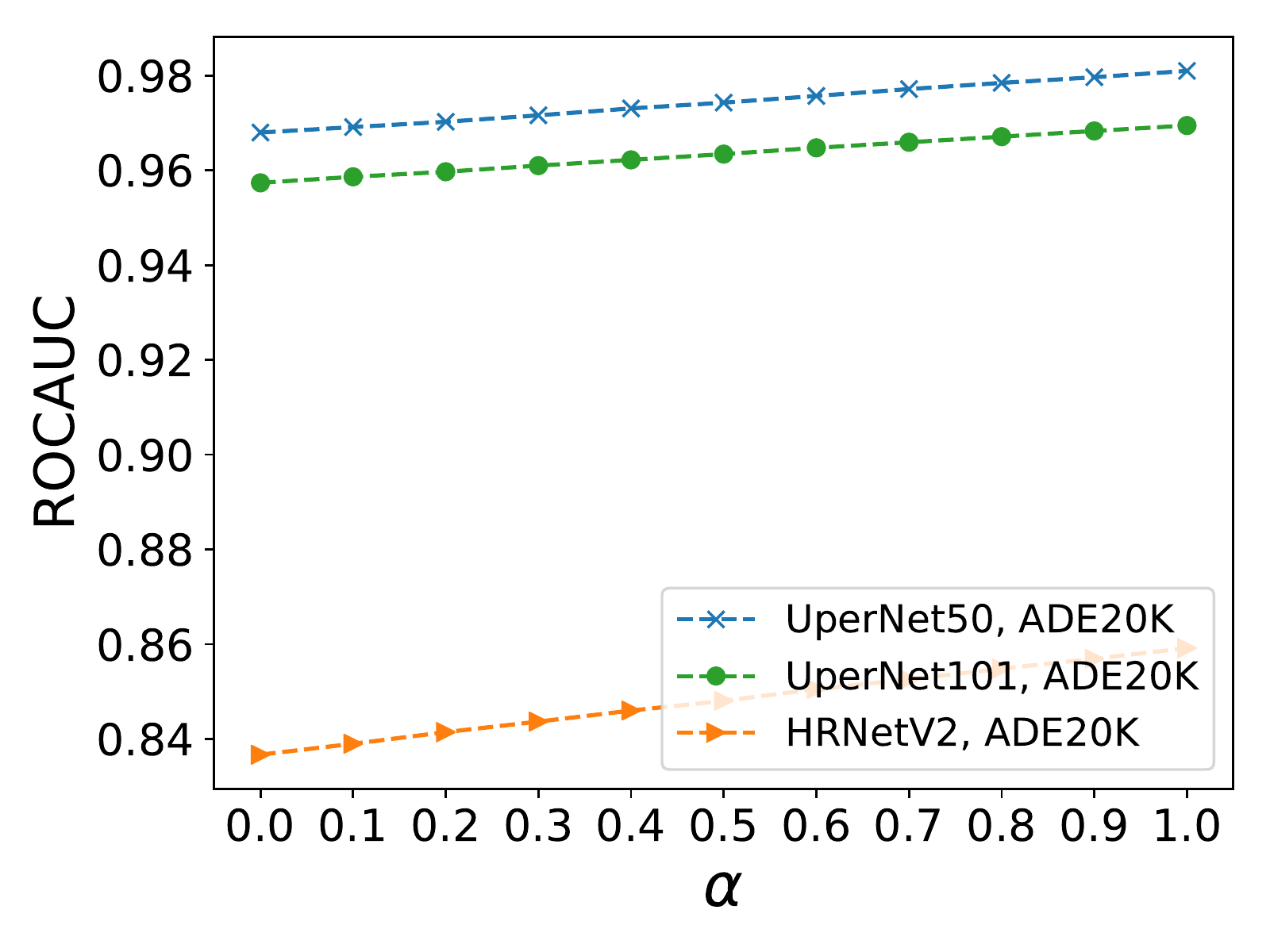}
}
\subfigure[SPADE]{
\includegraphics[width=0.46\linewidth, height=0.38\linewidth]{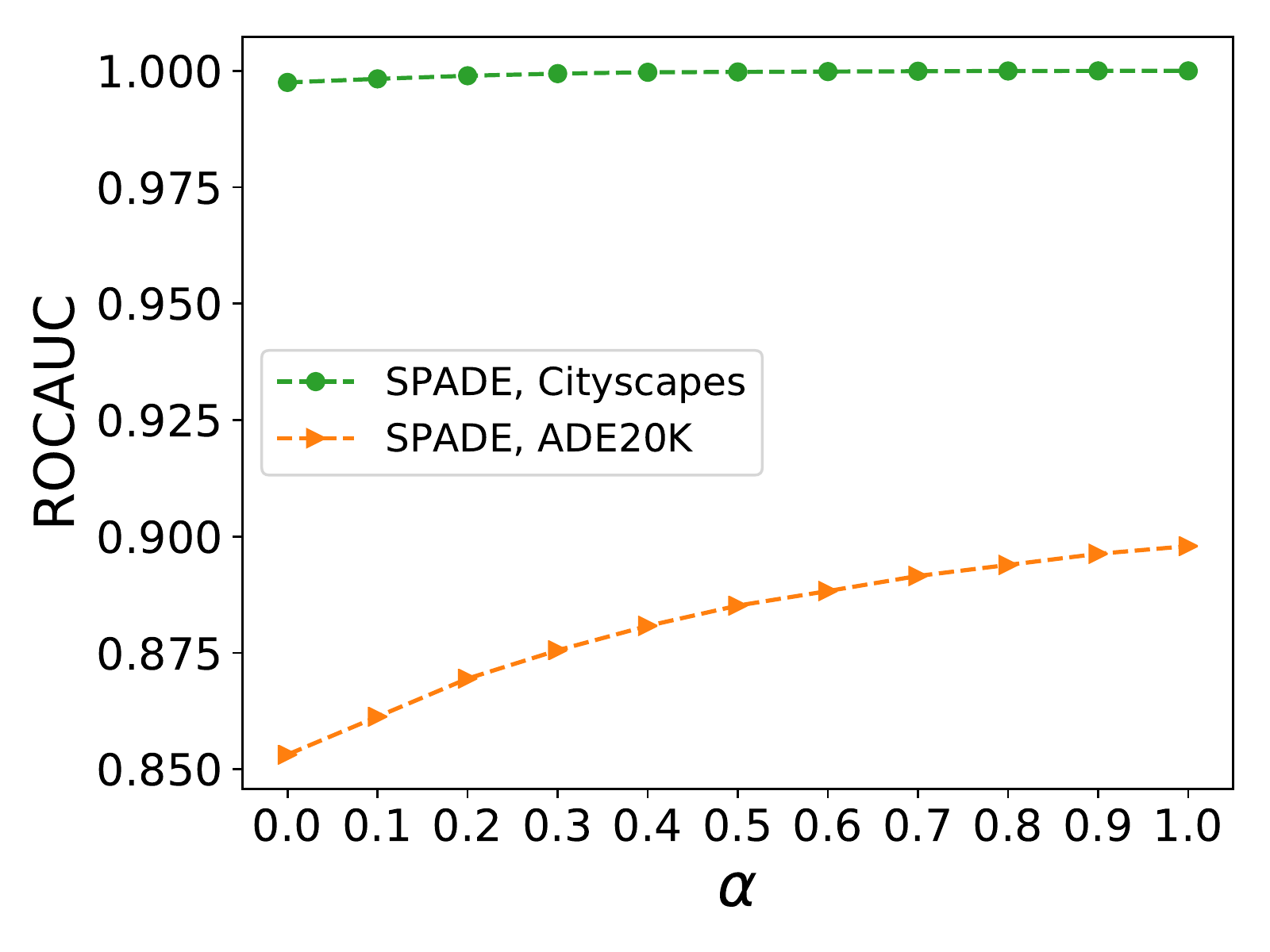}
}

\end{center}
\caption{Effect of $\alpha$ in Eq.~(\ref{eq:L-mem}) over the attack success.}
\label{fig:alpha_effect}
\end{figure}

We experimented with different values for the $\alpha$ value in Eq.~(\ref{eq:L-mem}). As can be seen in Fig.~\ref{fig:alpha_effect}, $\alpha = 1$ was the best choice over all benchmarks.

\begin{mdframed}
    \textbf{Algorithm 1. } Membership Inference Attack\\
    \textbf{Input:}  Query pair $(x, y)$, victim model $\mathbf{V}$, feature extractor $\mathbf{F}$, scalar $\alpha$, threshold $\tau$, error function $err$\\
    \textbf{Output:} Membership inference result\\
    \parbox{\textwidth}{\begin{enumerate}
        \item $L_{rec} = err(\mathbf{V}(x), y)$ 
        \item $\psi = \mathbf{F}(x) //$ $|\psi| = 56\times56\times3840$
        \item $y = resize(y, 56\times56\times3)$
        \item $\{\psi_{train}, y_{train}\} \xleftarrow{70\%} \{\psi, y\}$ 
        \item $\{\psi_{test}, y_{test}\} = \{\psi, y\} \setminus \{\psi_{train}, y_{train}\}$ 
        \item Train linear regression $\mathbf{P}$ with $\{\psi_{train}, y_{train}\}$
        \item $L_{pred} = \frac{1}{N}\sum_{i=1}^{N} \| \mathbf{P} \psi_{test} (i) - y_{test}^i\| _1$ $// N = 941$
        \item $L_{mem} = L_{rec} - \alpha \cdot L_{pred}$
        \item \textbf{if} $L_{mem} < \tau$ \textbf{then}\\
        \hspace*{1em} Return \textbf{True} \\ 
        \textbf{else}\\
        \hspace*{1em} Return \textbf{False} \\
    \end{enumerate}}%
\label{alg:attack}
\end{mdframed}

\subsection{MIA vs output dimension}

\begin{figure}[t]
\begin{center}
\subfigure[Pix2pix]{
\includegraphics[width=0.46\linewidth, height=0.38\linewidth]{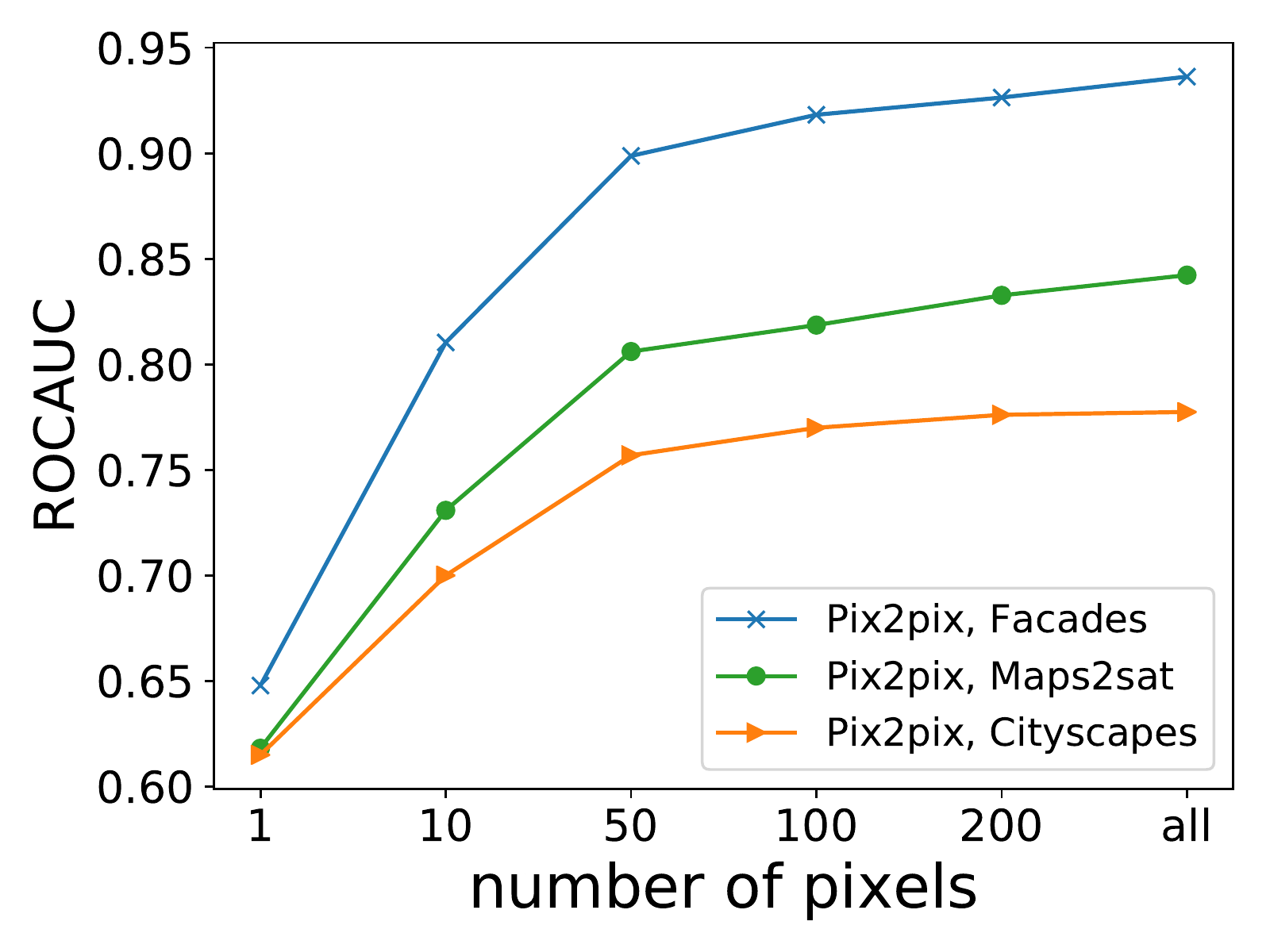}
}
\subfigure[SPADE]{
\includegraphics[width=0.46\linewidth, height=0.38\linewidth]{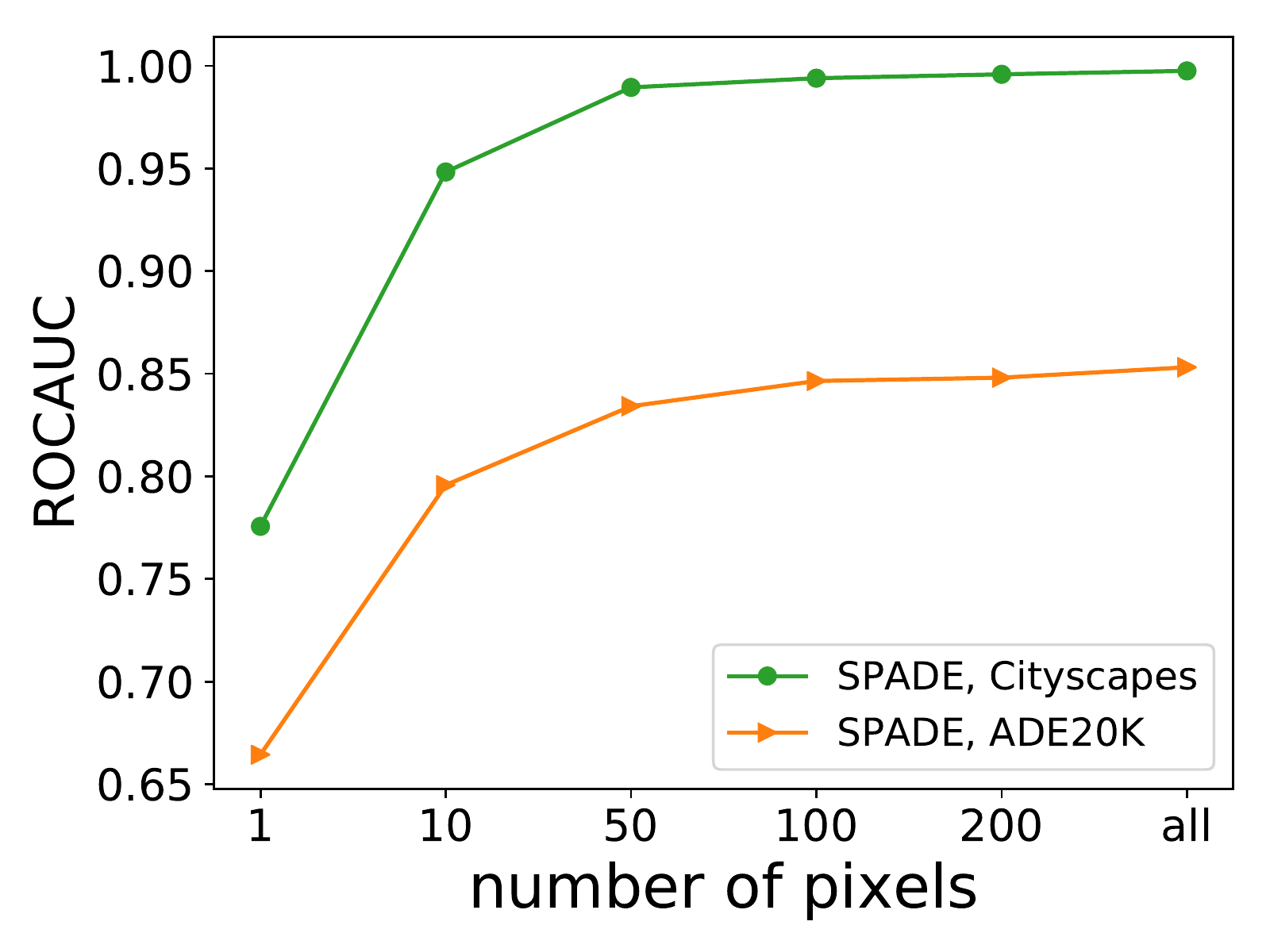}
}
\end{center}
\caption{Effect of reducing output dimensionality over a reconstruction-based attack. MIA accuracy is correlated with the decrease of output dimension, i.e.\ number of pixels, demonstrating that high output dimensionality problems are more vulnerable to MIA.}
\label{fig:rest_dim_reduce_effect}
\end{figure}

As described in Sec.~\ref{sec:investigation_experimetns}, we evaluted the effect of reducing the output dimension on the accuracy of reconstruction-based MIA. The reduction was achieved by randomly sampling $N$ output pixels, and using them as the output, where $N$ ranges from a single pixel and up to $200$ pixels. Fig.~\ref{fig:rest_dim_reduce_effect} demonstrates that MIA accuracy indeed scales with the number of output dimensions. Results for Pix2PixHD, UperNet and HRNetV2 are presetned in Fig.~\ref{fig:dim_reduce_effect}.

\subsection{calibration Effect}
\begin{figure*}[tb]
\begin{center}
\subfigure[Pix2pixHD - Facades]{
\includegraphics[width=0.3\linewidth, height=0.2\linewidth]{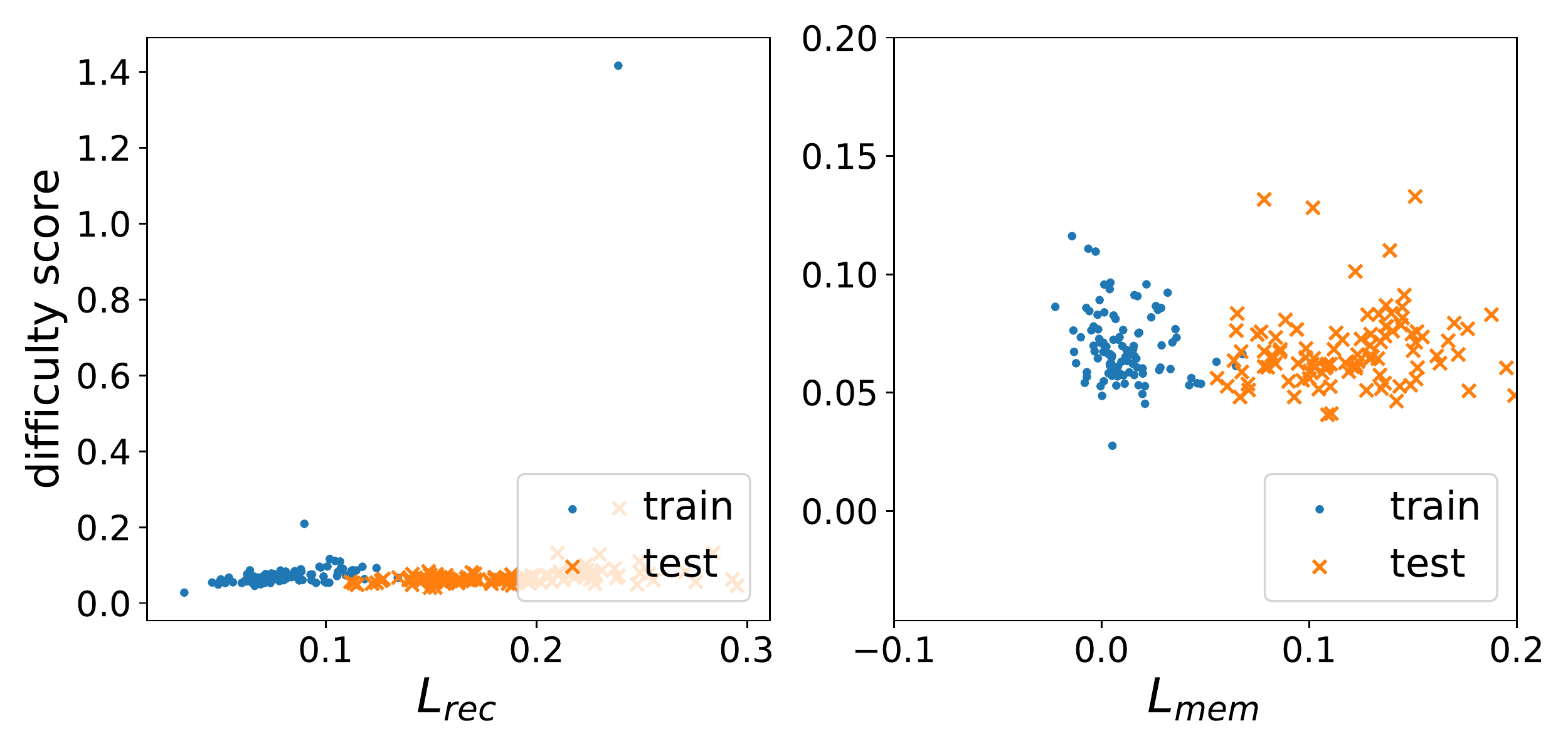}
}
\subfigure[Pix2pixHD - CelebA]{
\includegraphics[width=0.3\linewidth, height=0.2\linewidth]{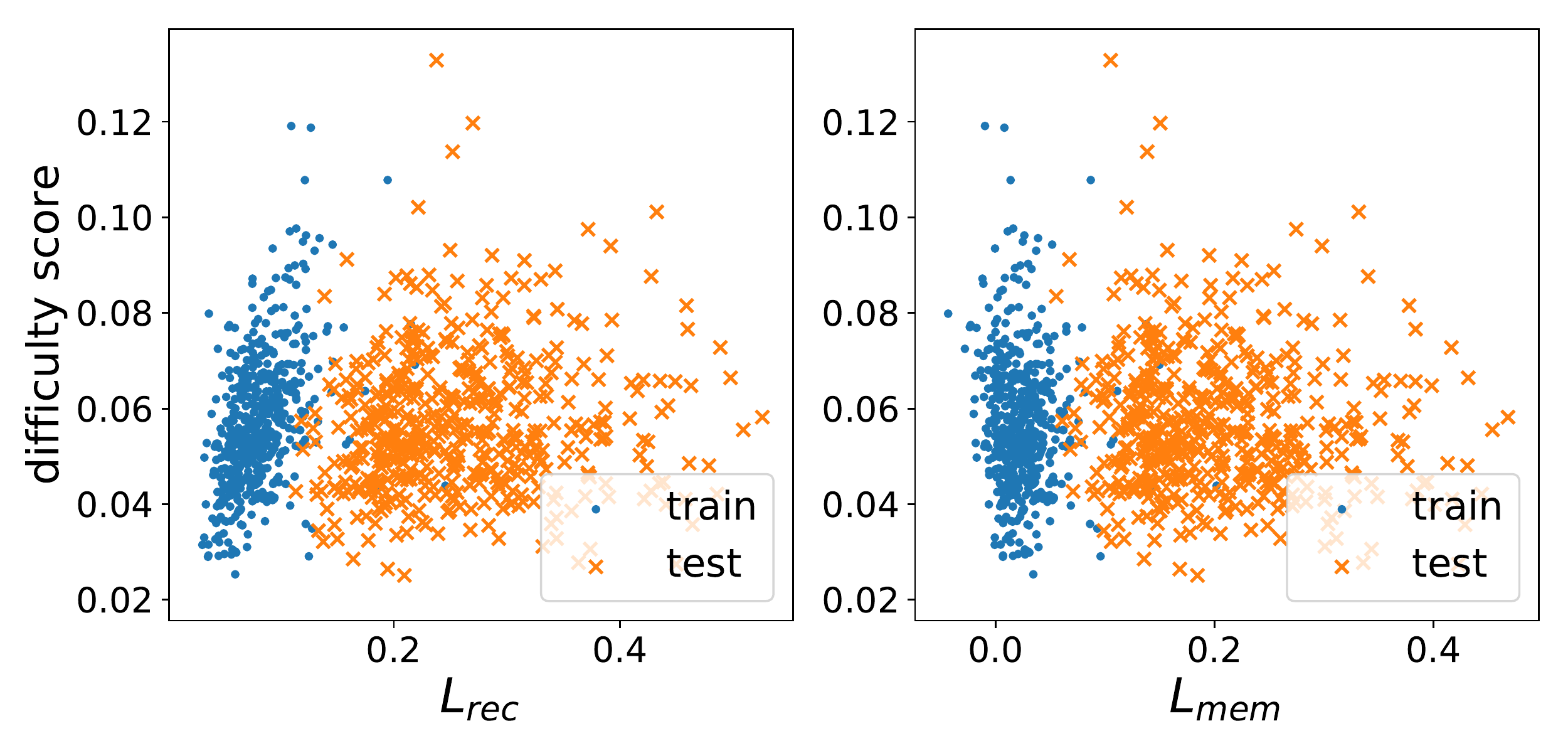}
}\\
\subfigure[Pix2pix - Facades]{
\includegraphics[width=0.3\linewidth, height=0.2\linewidth]{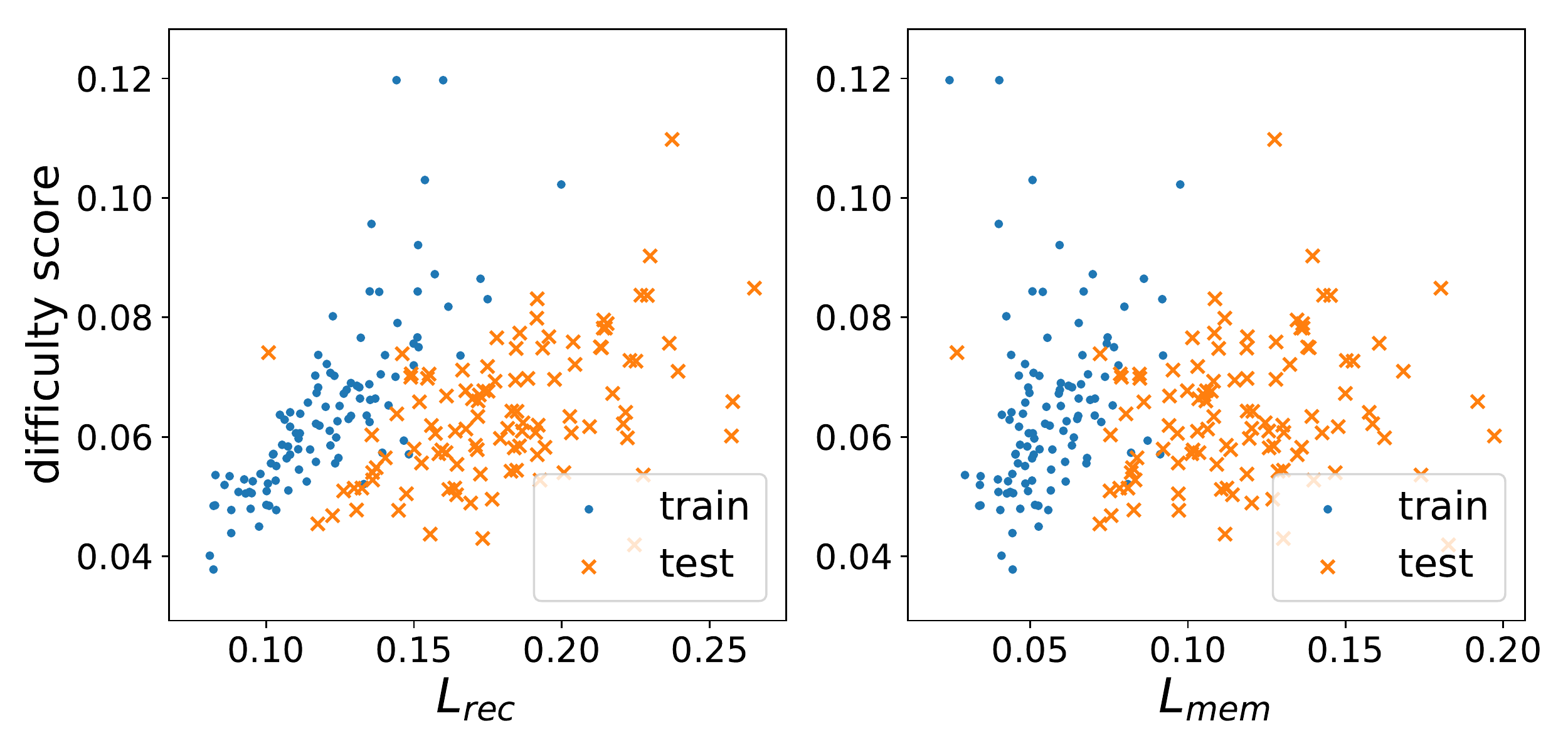}
}
\subfigure[Pix2pix - Maps]{
\includegraphics[width=0.3\linewidth, height=0.2\linewidth]{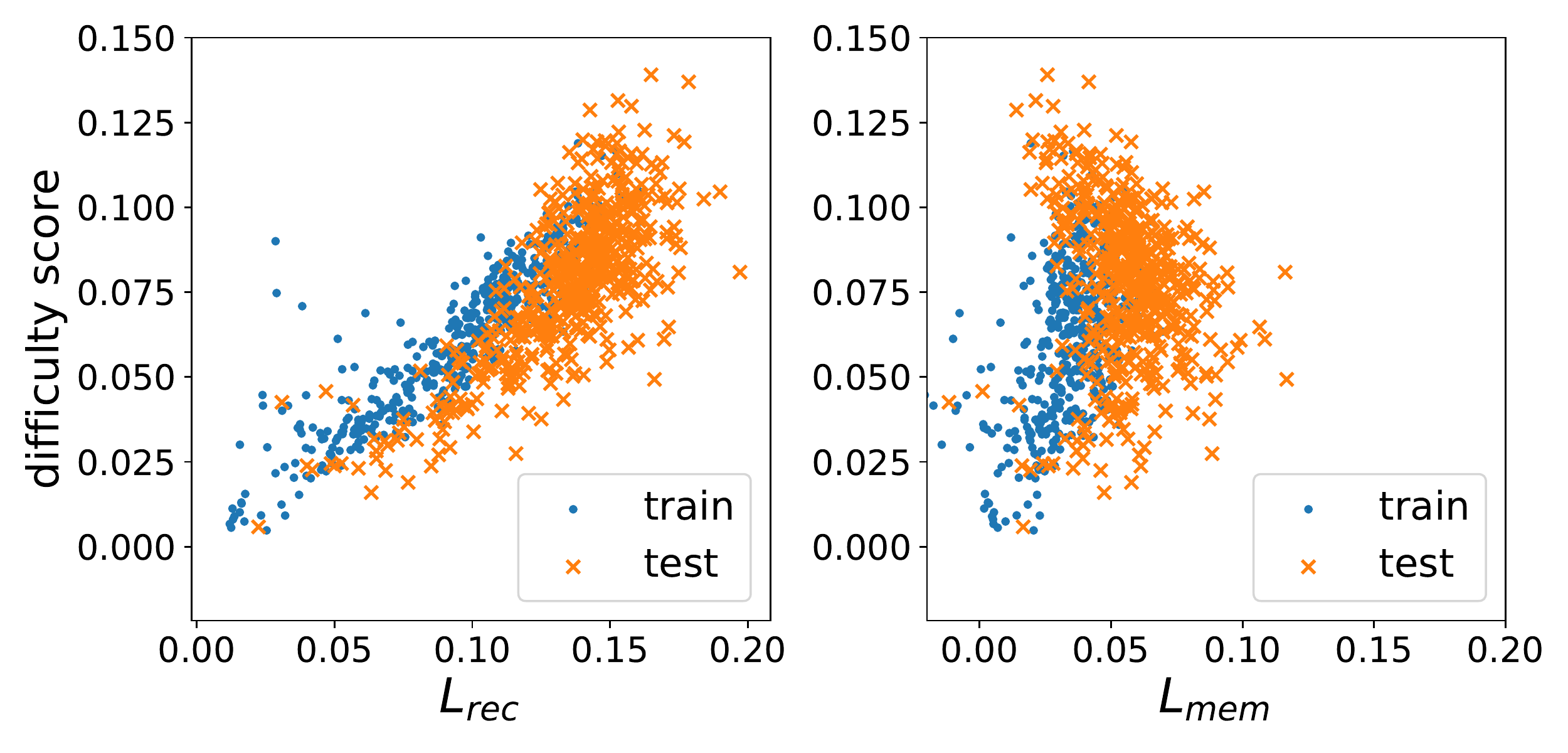}
}
\subfigure[Pix2pix - Cityscapes]{
\includegraphics[width=0.3\linewidth, height=0.2\linewidth]{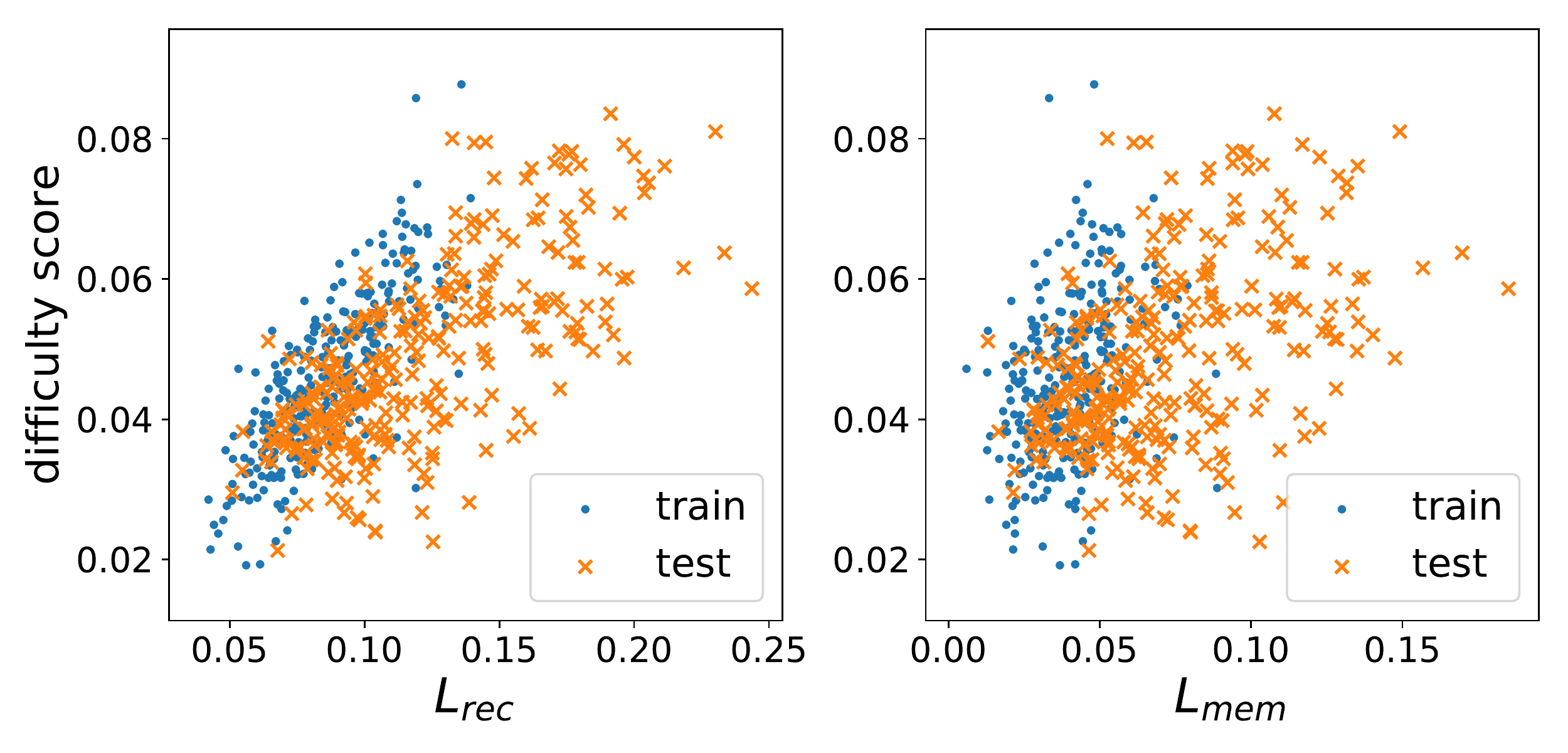}
}\\
\subfigure[SPADE - Cityscapes]{
\includegraphics[width=0.3\linewidth, height=0.2\linewidth]{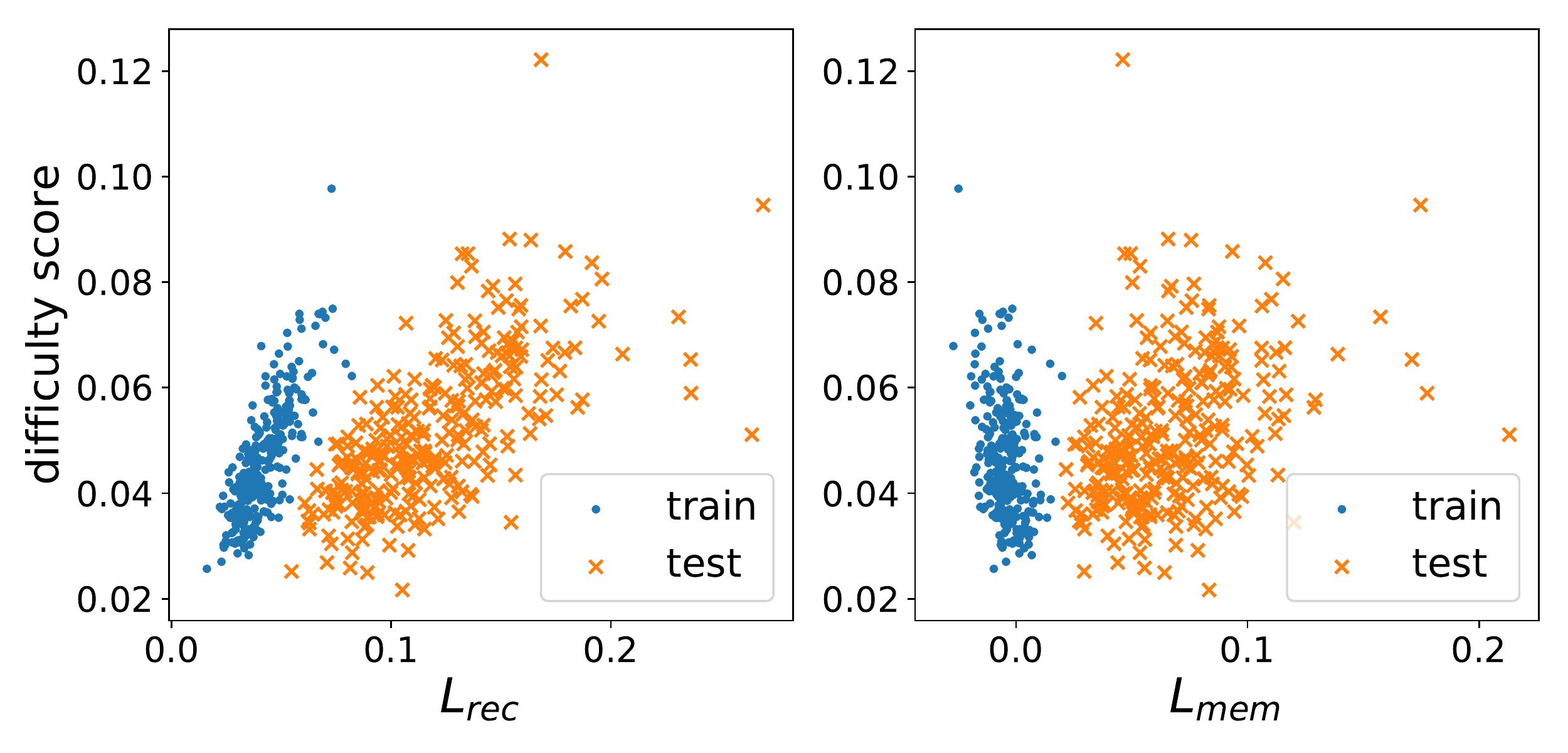}
}
\subfigure[SPADE - ADE20K]{
\includegraphics[width=0.3\linewidth, height=0.2\linewidth]{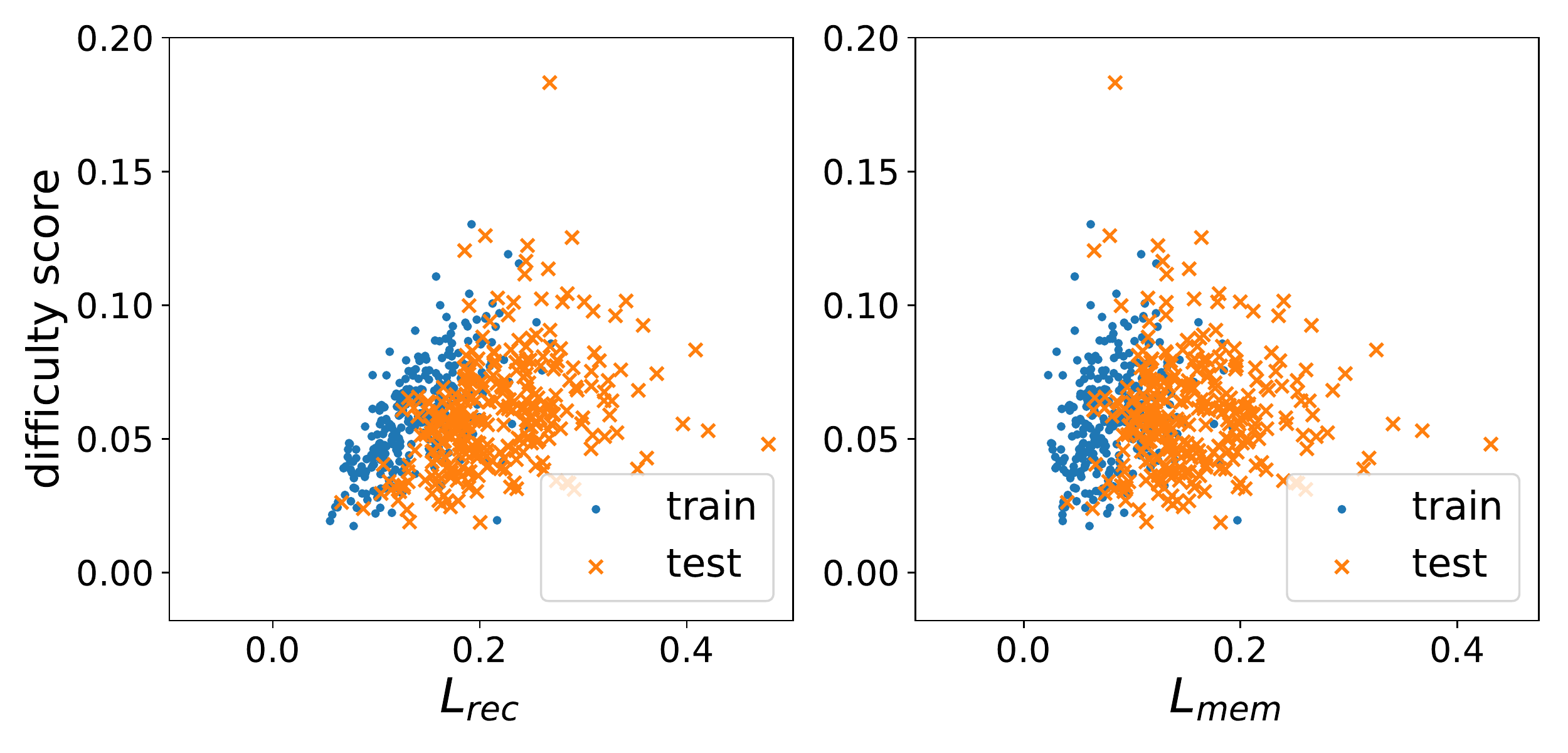}
}\\
\subfigure[UperNet50 - ADE20K]{
\includegraphics[width=0.3\linewidth, height=0.2\linewidth]{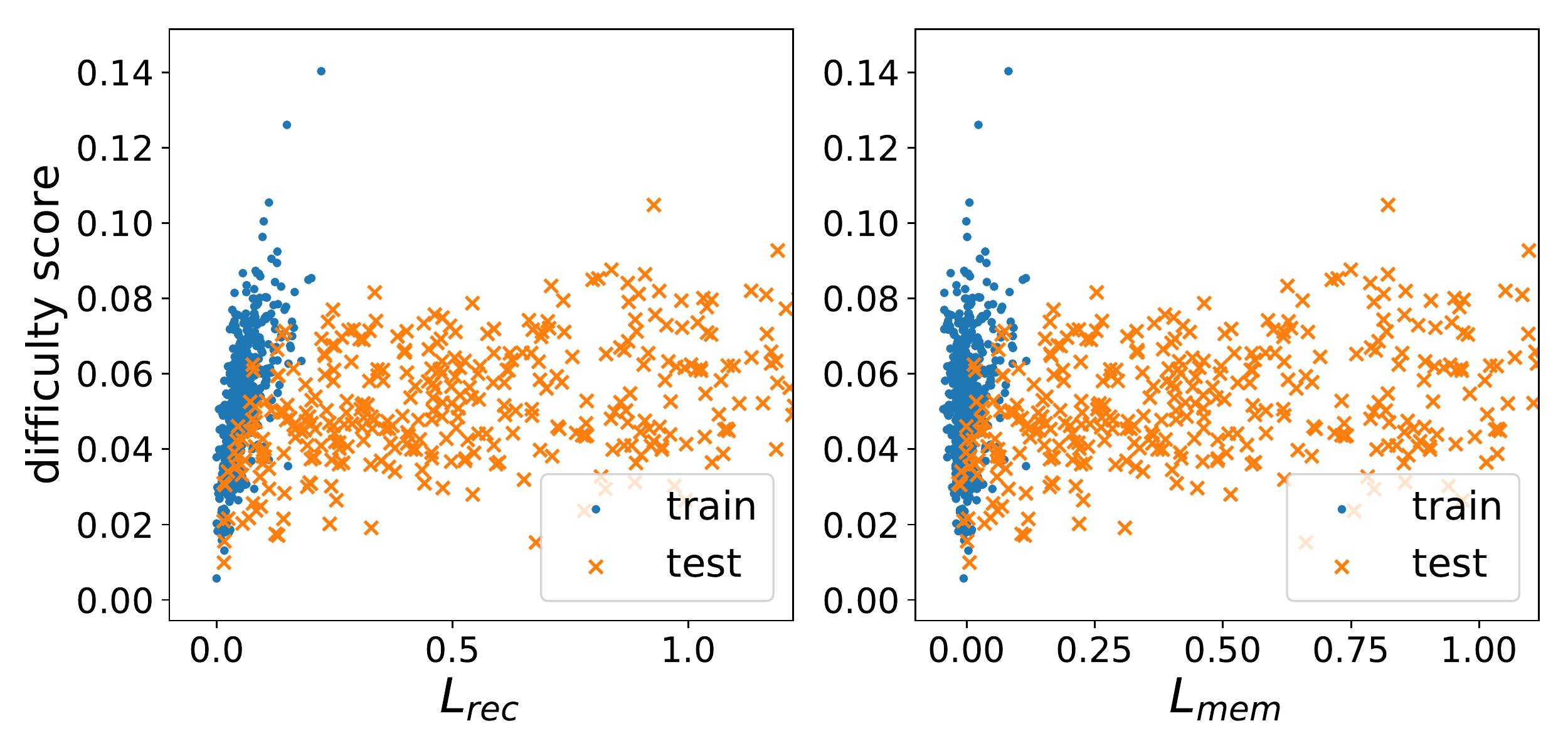}
}
\subfigure[UperNet101 - ADE20K]{
\includegraphics[width=0.3\linewidth, height=0.2\linewidth]{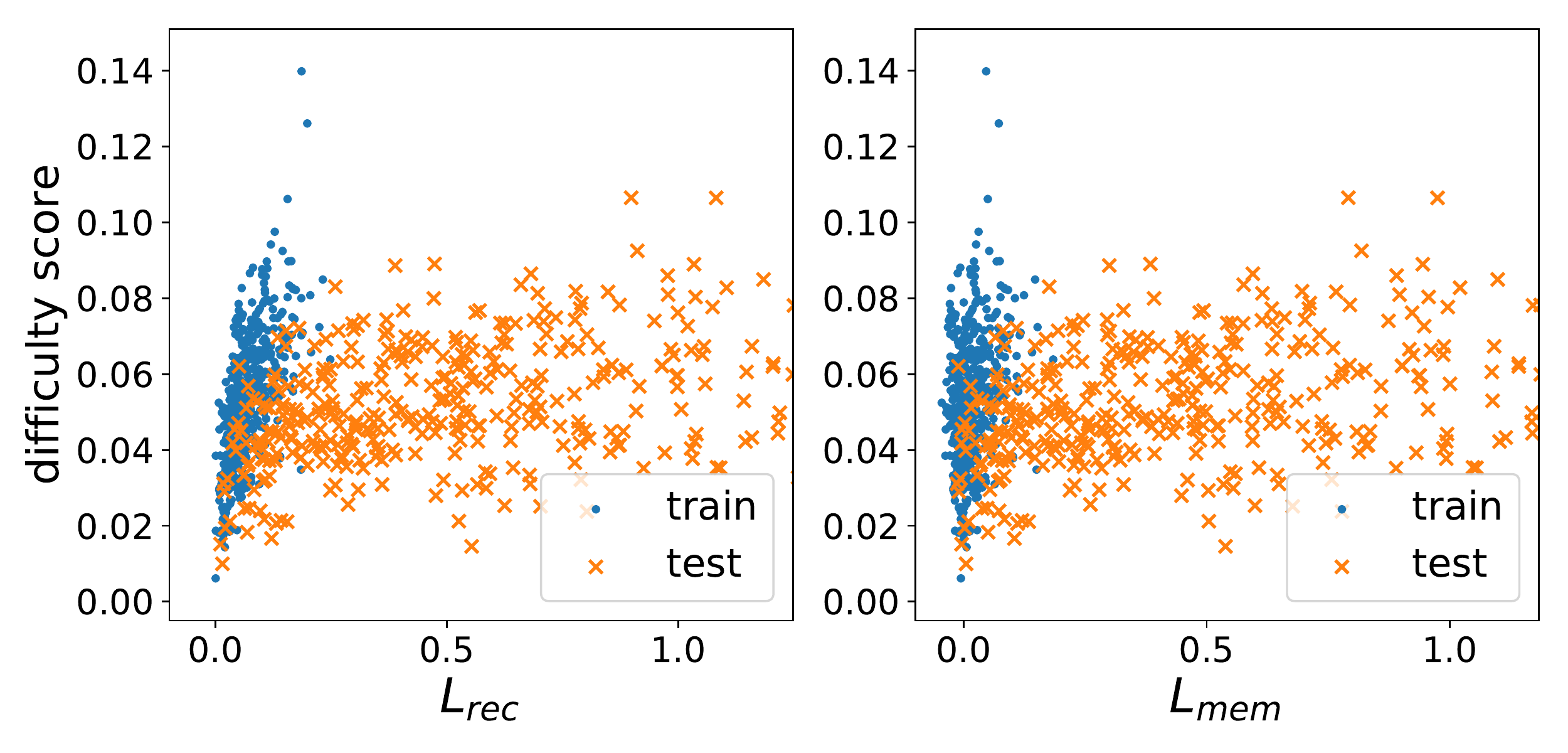}
}
\subfigure[HRNetV2 - ADE20K]{
\includegraphics[width=0.3\linewidth, height=0.2\linewidth]{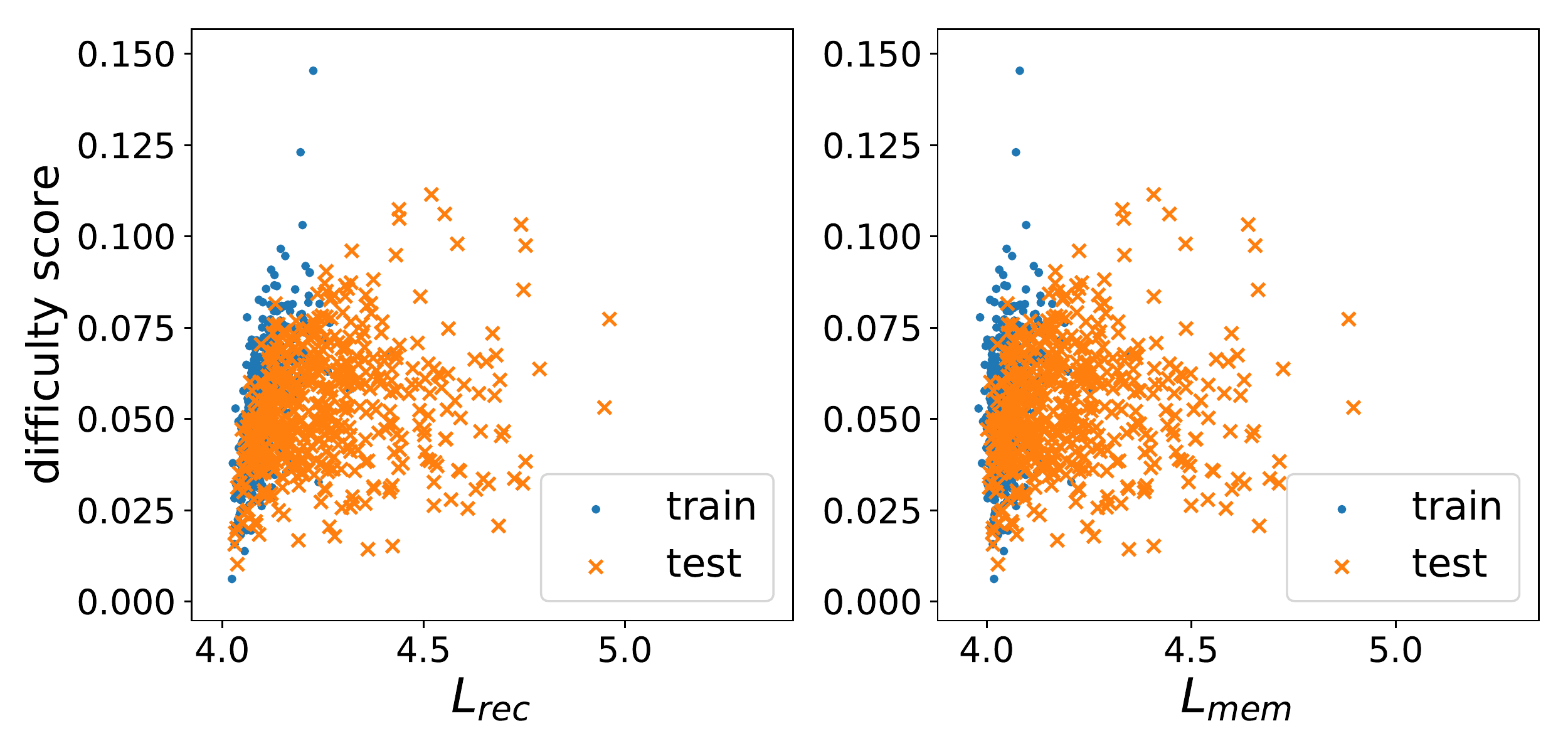}
}
\end{center}
\caption{The proposed membership error $L_{mem}$ can better separate train and test images by a simple threshold (i.e. a vertical line) compared to the reconstruction error $L_{rec}$. Pix2pixHD for Maps2sat and Cityscapes are presented in Fig.~\ref{fig:calibration_effect}}
\label{fig:rest_of_calibration_effect}

\end{figure*}

As can be seen in Tab.~\ref{tab:main_method_results}, using our membership error $L_{mem}$, Eq.~(\ref{eq:L-mem}), substantially improves the sucess rates in all of our experiments. As can be seen in Fig.~\ref{fig:rest_of_calibration_effect}, our $L_{mem}$ can better separate train and test images by a simple threshold compared to the reconstruction error $L_{rec}$. Results for Pix2PixHD on the Maps2sat and Cityscapes datasets are presented in Fig.~\ref{fig:calibration_effect}.

\subsection{Human-Supervised Image difficulty score}
\label{app:supervised_diff_score}
\begin{figure}[h]
\begin{center}
\includegraphics[width=0.15\linewidth, height=0.18\linewidth]{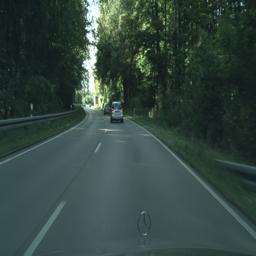}
\includegraphics[width=0.15\linewidth, height=0.18\linewidth]{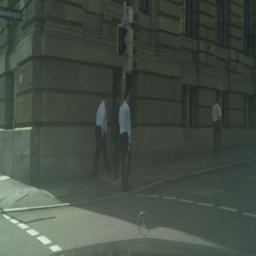}
\includegraphics[width=0.15\linewidth, height=0.18\linewidth]{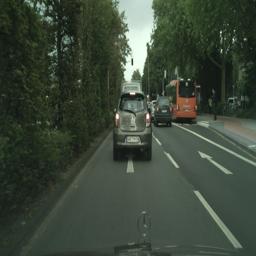}
\includegraphics[width=0.15\linewidth, height=0.18\linewidth]{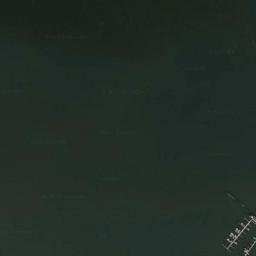}
\includegraphics[width=0.15\linewidth, height=0.18\linewidth]{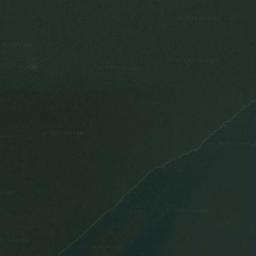}
\includegraphics[width=0.15\linewidth, height=0.18\linewidth]{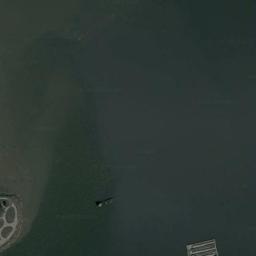}\\
\includegraphics[width=0.15\linewidth, height=0.18\linewidth]{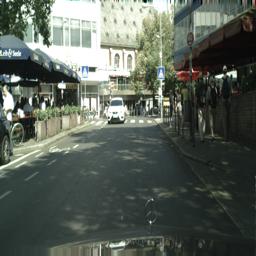}
\includegraphics[width=0.15\linewidth, height=0.18\linewidth]{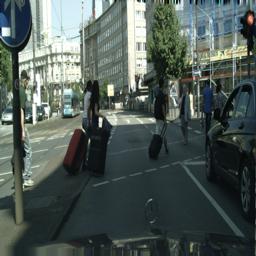}
\includegraphics[width=0.15\linewidth, height=0.18\linewidth]{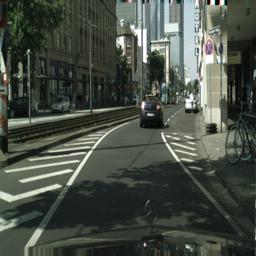}
\includegraphics[width=0.15\linewidth, height=0.18\linewidth]{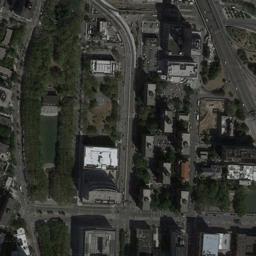}
\includegraphics[width=0.15\linewidth, height=0.18\linewidth]{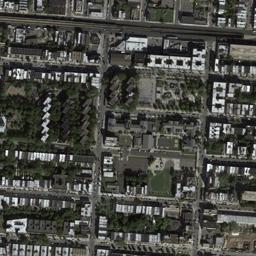}
\includegraphics[width=0.15\linewidth, height=0.18\linewidth]{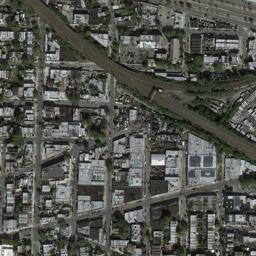}

\end{center}
\caption{Examples of images from the Cityscapes (first two rows) and Maps2sat (last two rows) datasets that received the lowest (first and third row) and highest (second and last row) predictability errors using the supervised difficulty score.}
\label{fig:supervised_samples}
\end{figure}

We compare our self-supervised single-sample predictability error with the human-supervised difficulty score proposed by \cite{tudor2016hard}.
In Fig.~\ref{fig:supervised_samples}, we present images ranked from easy to difficult using our implementation of the supervised-image difficulty score, for the Cityscapes and Maps datasets. The ranking seems correlated with image sharpness and level-of-detail images. 
As can be seen in Tab.~\ref{tab:single_vs_multi_diff_score}, our score outperforms the human-supervised score. We compare the correlation between the reconstruction error for unseen images to our self-supervised predictability error and the human-supervsied scorre.

\begin{table}[h]
\centering
\small

    \begin{tabular}{lccc}
    \toprule
    \textbf{Model} & \textbf{Dataset} & \textbf{Ours} & \textbf{Human-Supervised}\\
    \textbf{} & \textbf{} & train / test & train / test\\

    \midrule
    Pix2Pix & Facades & 0.79 / 0.50 & -0.02 / 0.16 \\
    Pix2Pix & Maps2sat & 0.51 / 0.77 & 0.79 / 0.52 \\
    Pix2Pix & Cityscapes & 0.78 / 0.71 & 0.04 / 0.09\\
    \midrule
    Pix2PixHD & Facades & 0.67 / 0.36 &  0.27 / 0.04 \\
    Pix2PixHD & Maps2sat & 0.38 / 0.79 &  0.77 / 0.56 \\
    Pix2PixHD & Cityscapes & 0.76 / 0.62 &  0.36 / 0.48 \\
    \midrule
    SPADE & Cityscapes & 0.80 / 0.68 & 0.29 / 0.53 \\
    SPADE & ADE20K & 0.48 / 0.27 & 0.25 / -0.05 \\
    \midrule
    UperNet50 & ADE20K & 0.66 / 0.13 & 0.34 / 0.05 \\
    UperNet101 & ADE20K & 0.65 / 0.13 & 0.38 / 0.06  \\
    HRNetV2 & ADE20K & 0.61 / 0.22 &  0.36 / 0.10 \\
	\bottomrule
    \end{tabular}
    
\caption{Our self-supervised difficulty score is better correlated with the reconstruction error than the human-supervised}

\label{tab:supervised_score_correlation}
\end{table}

\subsection{Multi-Image predictability error}
\label{app:multi_image}
\begin{figure}[tb]
\begin{center}
\subfigure[Pix2pix-Maps2sat]{
\includegraphics[width=0.45\linewidth, height=0.4\linewidth]{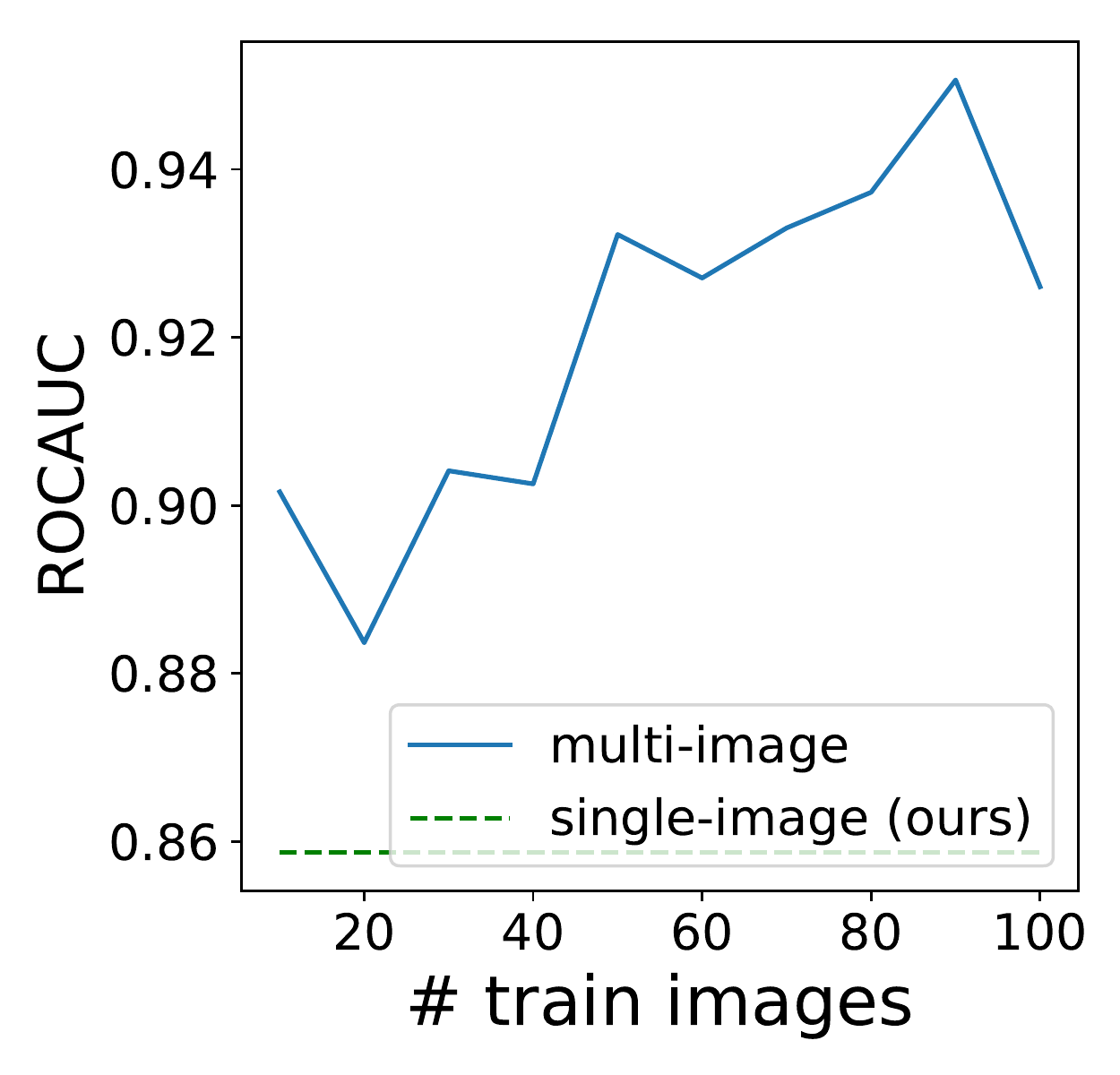}
}
\subfigure[Pix2pix-Cityscapes]{
\includegraphics[width=0.45\linewidth, height=0.4\linewidth]{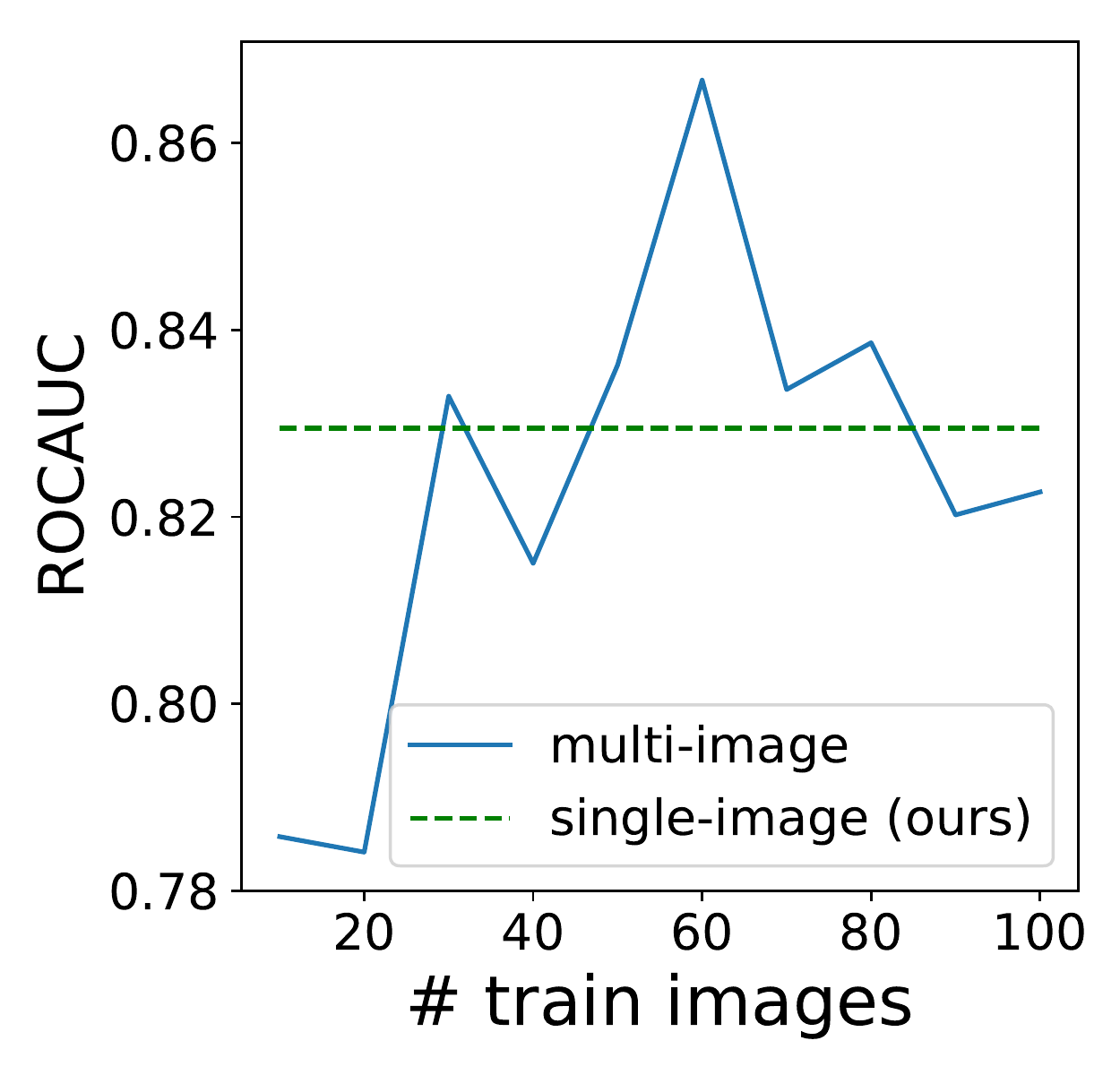}
}\\
\subfigure[Pix2pixHD-Maps2sat]{
\includegraphics[width=0.45\linewidth, height=0.4\linewidth]{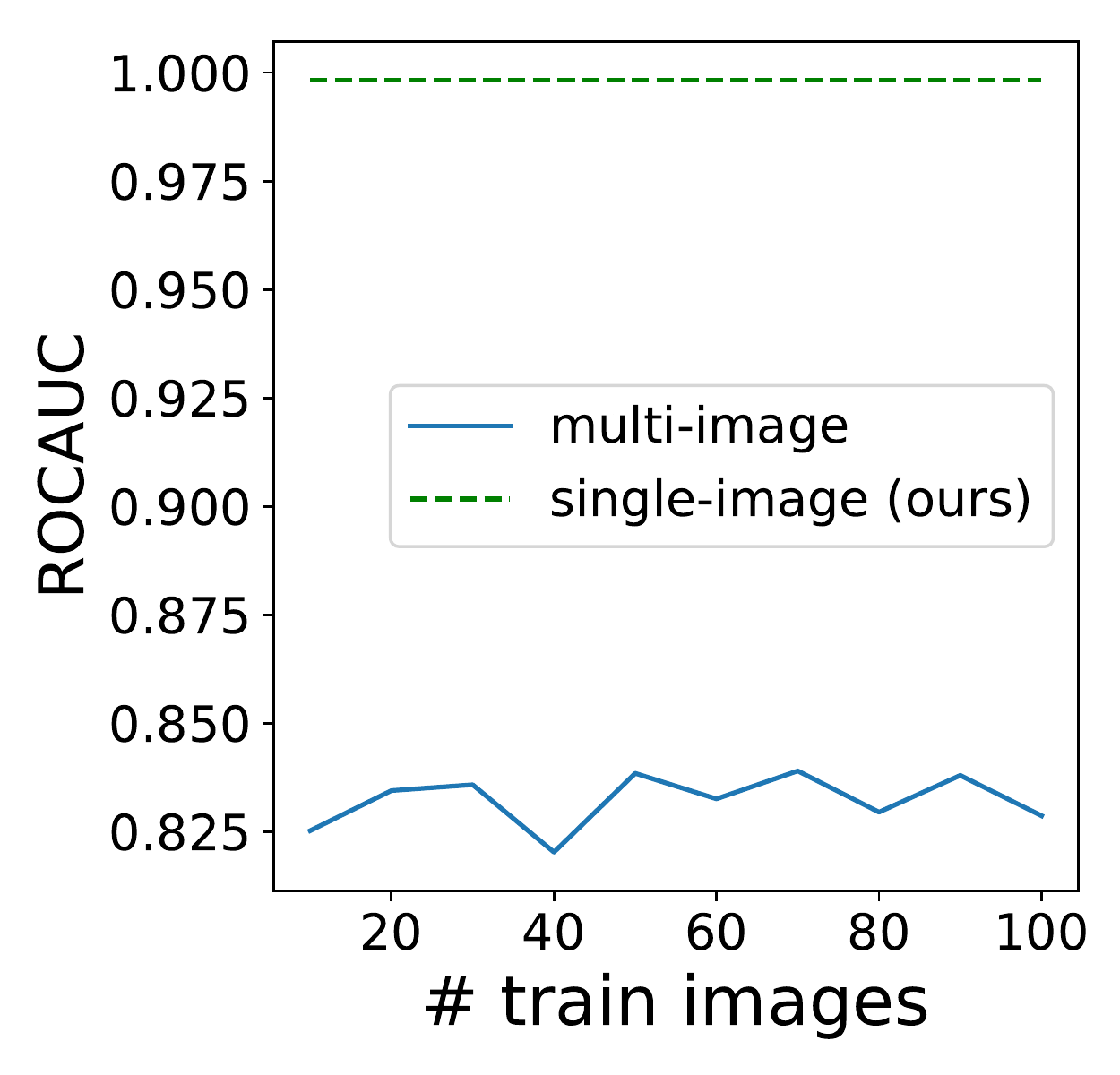}
}
\subfigure[Pix2pixHD-Cityscapes]{
\includegraphics[width=0.45\linewidth, height=0.4\linewidth]{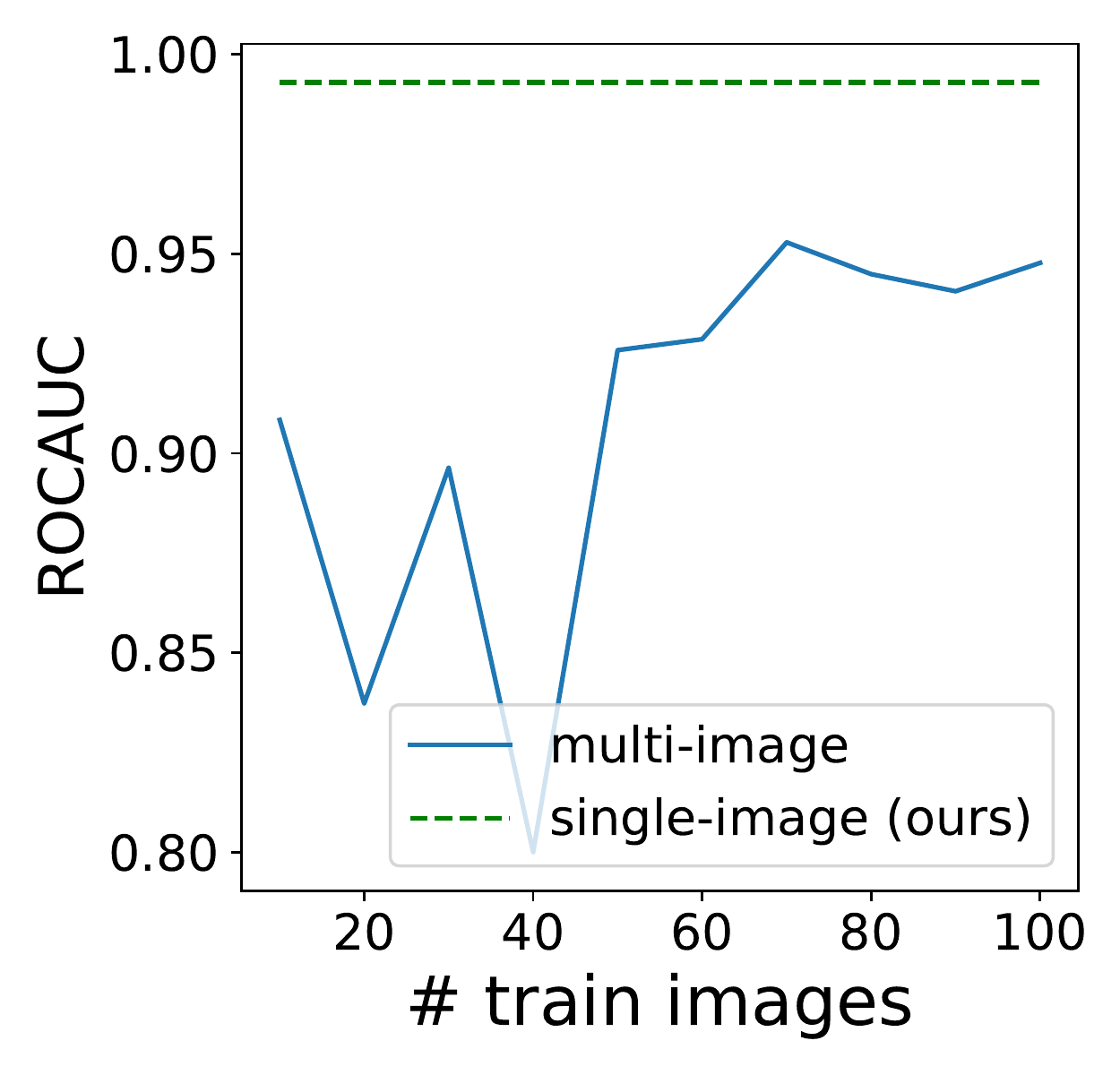}
}\\
\subfigure[UperNet50-ADE20K]{
\includegraphics[width=0.45\linewidth, height=0.4\linewidth]{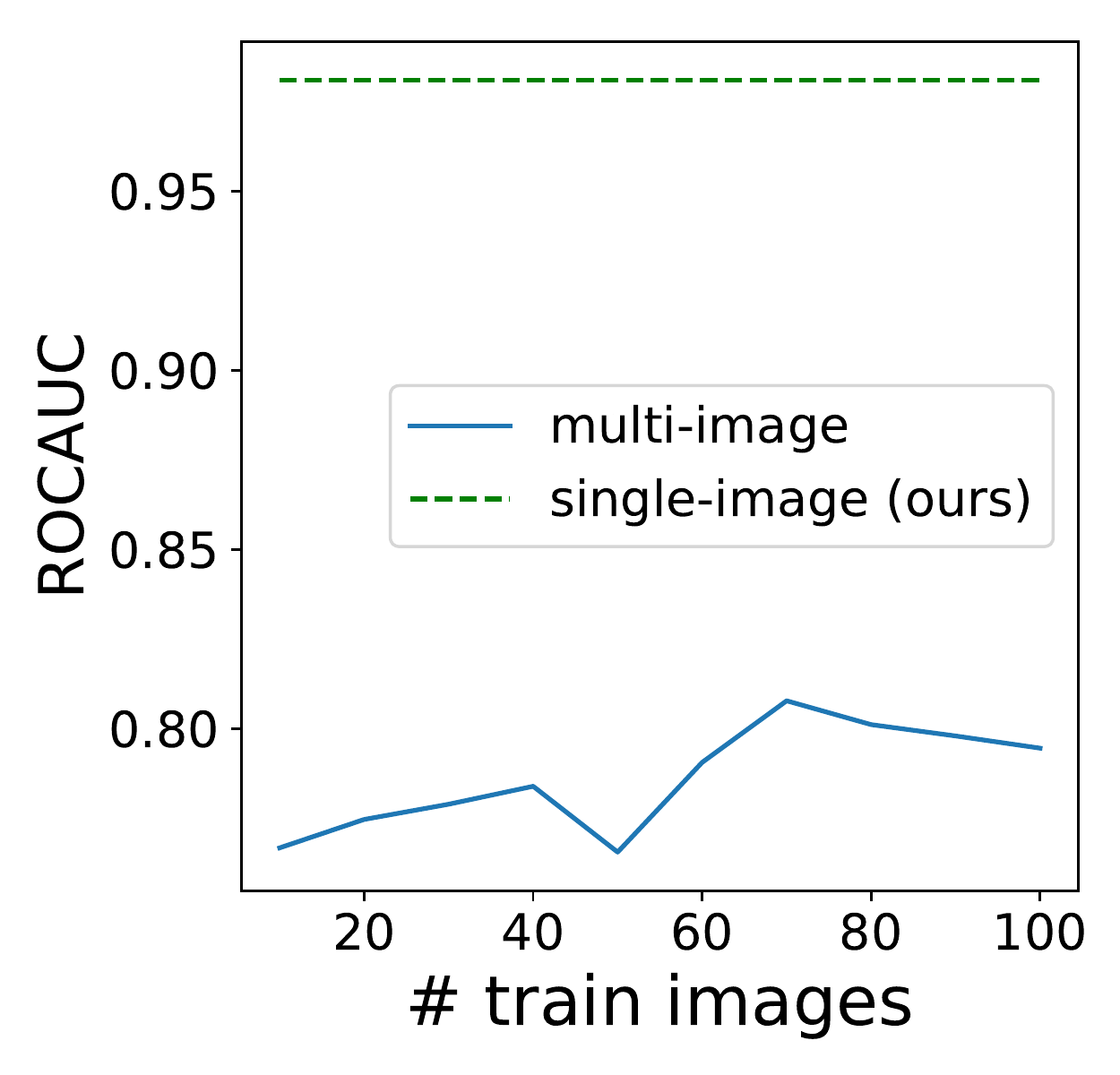}
}
\subfigure[UperNet101-ADE20K]{
\includegraphics[width=0.45\linewidth, height=0.4\linewidth]{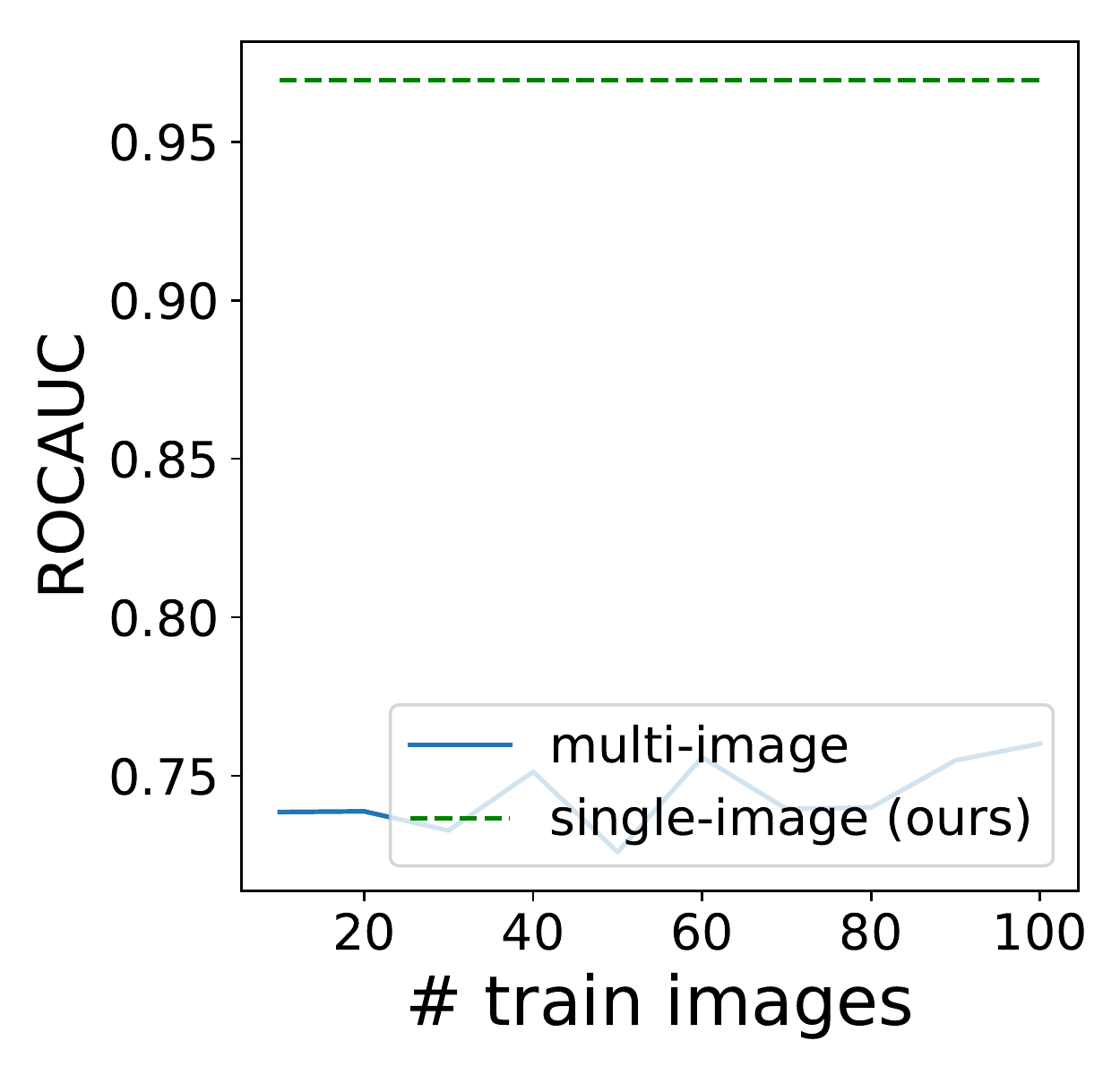}
}\\
\subfigure[HRNetV2-ADE20K]{
\includegraphics[width=0.45\linewidth, height=0.4\linewidth]{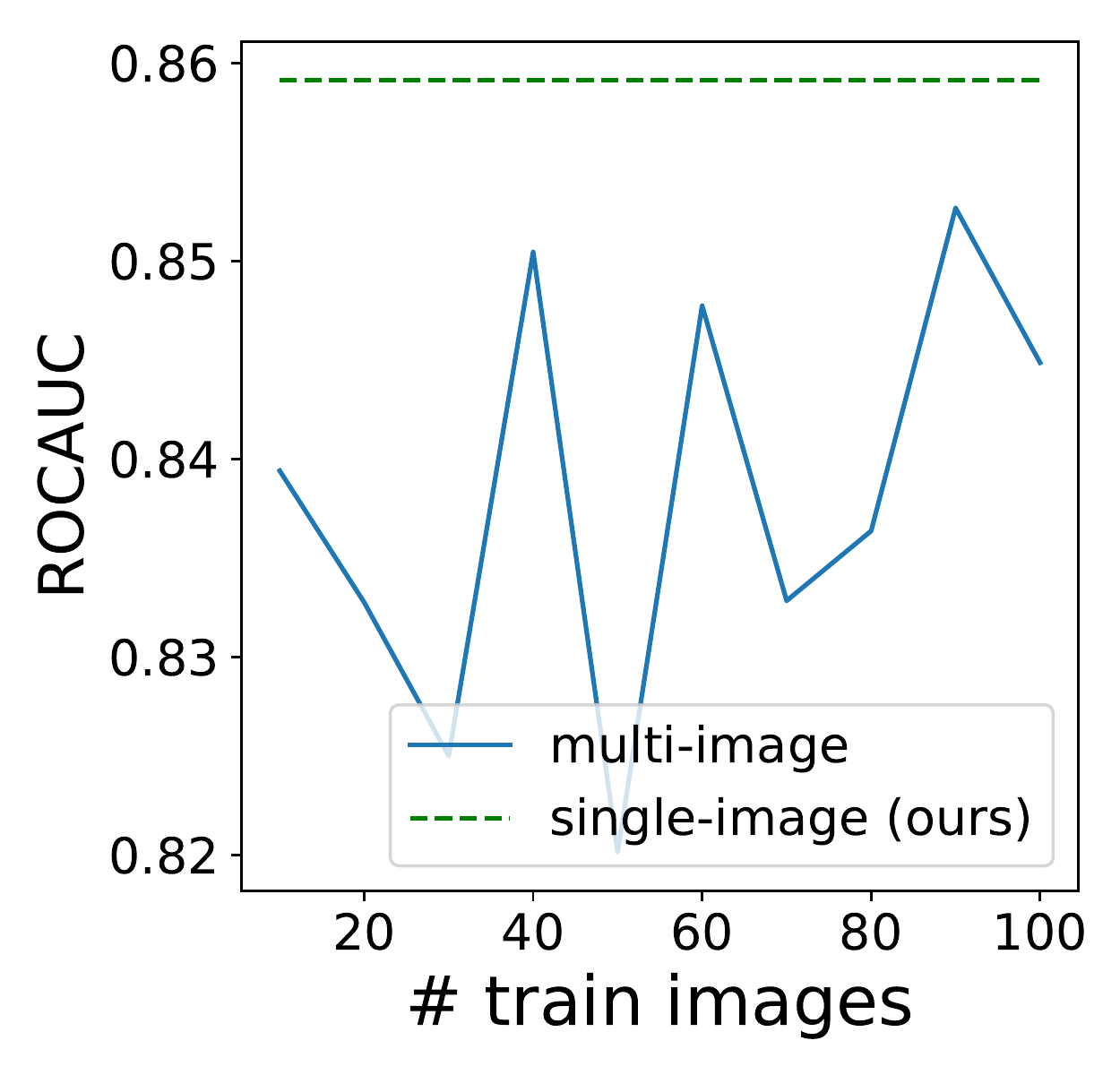}
}

\end{center}
\caption{Comparison of MIA accuracy when using our single sample vs. using multi-sample predictability errors, as a function of the number of training images. Note that the multi-image score assumes knowledge of the victim's model, as well as the availability of many labeled training images}
\label{fig:multi_image_comparison}

\end{figure}

As discussed in Sec.~\ref{sec:single_vs_multi_diff_score}, we compare our single-sample predictability error to a multi-sample predictability error (MSPS) by training a "shadow" model, sharing the same architecture as the victim model, on auxiliary samples. As can be seen in Tab.~\ref{tab:single_vs_multi_diff_score}, when training the MSPS on $100$ images, it underperforms our method on Pix2PixHD and the evaluated semantic segmentation models. For the smaller Pix2Pix architecture, MSPS was more successful, obtaining competitive results with our method. We analyzed the effect of number of samples over the MSPS performance. As can be seen in Fig.~\ref{fig:multi_image_comparison}, in most tasks, increasing the number of samples did not improve performance. 

We also compare our method to the setting were many out-of-distribution but similar sample are available. We trained shadow models on $4K$ samples from the BDD dataset as MSPS for the Cityscapes dataset. As can be seen in Tab.~\ref{tab:multi_image_bdd}, this too underperforms our method. Note that it is rare to have similar datasets with nearly identical labels, such as in the case of BDD and Cityscapes. 

\begin{table}[tb]
\centering
\small
    \begin{tabular}{lccccc}
    \toprule
    \textbf{Model} & \textbf{Dataset} & \textbf{Ours} &  \multicolumn{2}{c}{\textbf{Multi-Image}} \\
    \textbf{} & \textbf{} &  & In-Dist & BDD\\
    \midrule
    Pix2pix & Cityscapes & \textbf{82.94\%} & \textbf{82.47}\% & 74.43\%\\
    Pix2pixHD & Cityscapes & \textbf{99.29\%} & 96.86\% & 66.2\%\\
	\bottomrule
    \end{tabular}

\caption{Comparison between our single-image predictability error and two multi-image baselines, using in-distribution images and a larger amound of out-of-distribution images (BDD).}

\label{tab:multi_image_bdd}
\end{table}

\subsection{Shadow models}
\label{app:shadow_models}
\begin{table*}[tb]
\centering
\small
    \begin{tabular}{lcccccc}
    \toprule
    \textbf{Model} & \textbf{Dataset} & \textbf{Ours} & \multicolumn{2}{c}{\textbf{In-Dist}} & \multicolumn{2}{c}{\textbf{Out-of-Dist (BDD)}}\\
    \textbf{} & \textbf{} & ROC & ROC & Acc. & ROC & Acc.\\
    \midrule
    Pix2pix & Maps2sat & \textbf{85.65\%} & 80.15\% & 73.4\% & - & -\\
    Pix2pix & Cityscapes & \textbf{83.23\%} & 78.68\% & 67.5\% & 72.57\% & 56.16\% \\
    \midrule
    Pix2pixHD & Maps2sat & \textbf{99.42\%} & 98.63\% & 93.7\% & - & -\\
    Pix2pixHD & Cityscapes & \textbf{99.09\%} & 96.39\% & 64.0\% & 95.78\% & 56.5\%\\
	\bottomrule
    \end{tabular}

\caption{Comparison between our MIA and the commonly used shadow-model-based classifier attack, using $4K$ train and $4K$ test images from the BDD dataset. Our MIA outperforms while not requiring extra training images.}

\label{tab:shadow_model_bdd}
\end{table*}

As discussed in Sec.~\ref{sec:single_vs_multi_diff_score}, for the interest of completeness we compare our method with the popular approach of shadow-model classifiers for image translation MIA. For this, we select $N$ images, denoted as \textit{shadow\_train}, for training the shadow model. As an upper-bound, the shadow model is given the exactly same architecture as used by the victim model. Another $N$ images, not seen by the shadow model, are set to be \textit{shadow\_test}.
We define a new label for each sample as follows: 
\begin{equation}
        label(x)= 
\begin{cases}
    0,& \text{if } x\leftarrow shadow\_train\\
    1,              & \text{if } x\leftarrow shadow\_test
\end{cases}
\end{equation}
The classifier $\mathbf{C}$ architecture and training procedure are similar to \cite{he2019segmentations}. For each image, we compute the structured loss map between the ground-truth image and the generated image, using $L_1$ as the loss function. At every epoch we randomly crop $15$ patches of size $90\times90$ from the structured loss map. We train a ResNet-50 \cite{he2016deep} from scratch on the $90\times90$ patches, modified for binary classification.  We use a batch size of $8$, SGD optimizer, weight decay of $1e - 2$, initial learning rate of $0.1$ which reduces by a factor of $0.1$ every $15$ epochs. As previously mentioned, we do not evaluate this on the Facades dataset, due to its size.

We compare the performance of our single-sample method to the shadow model method in Tab.~\ref{tab:shadow_model_bdd}. For fairness we compare both the ROCAUC over the classifier's confidence, as well as the classification accuracy. It can be seen that in both cases, and for either in-distribution or out-of-distribution auxiliary data, the shadow model approach is inferior to our method for image translation models. We discuss the case of semantic segmentation in Sec.~\ref{sec:single_vs_multi_diff_score}.

\subsection{Memorization}
\label{app:overfittng}

\begin{figure}[h]
\begin{center}
\subfigure[Pix2pix]{
\includegraphics[width=0.4\linewidth, height=0.35\linewidth]{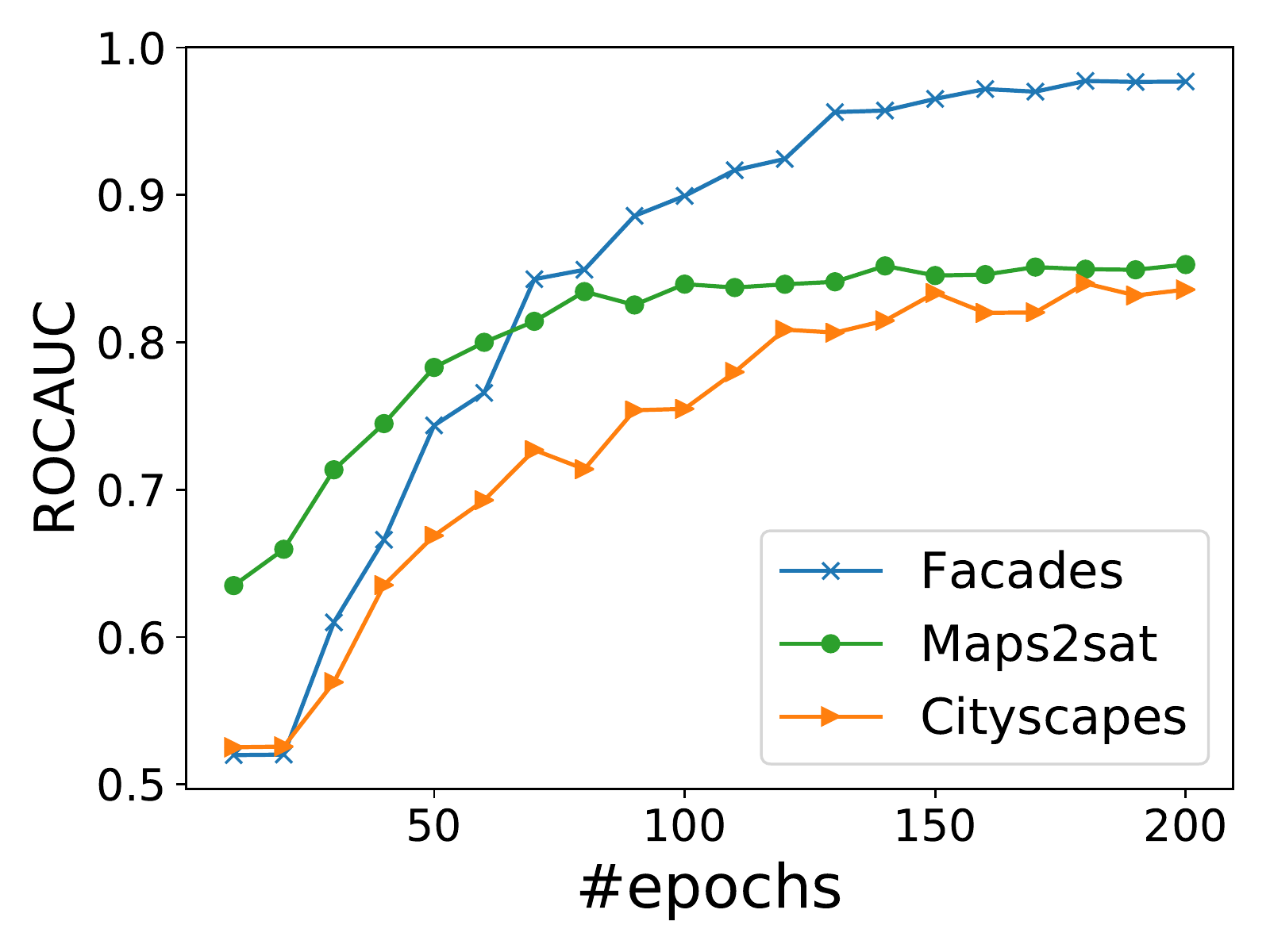}
}
\subfigure[Pix2pixHD]{
\includegraphics[width=0.4\linewidth, height=0.35\linewidth]{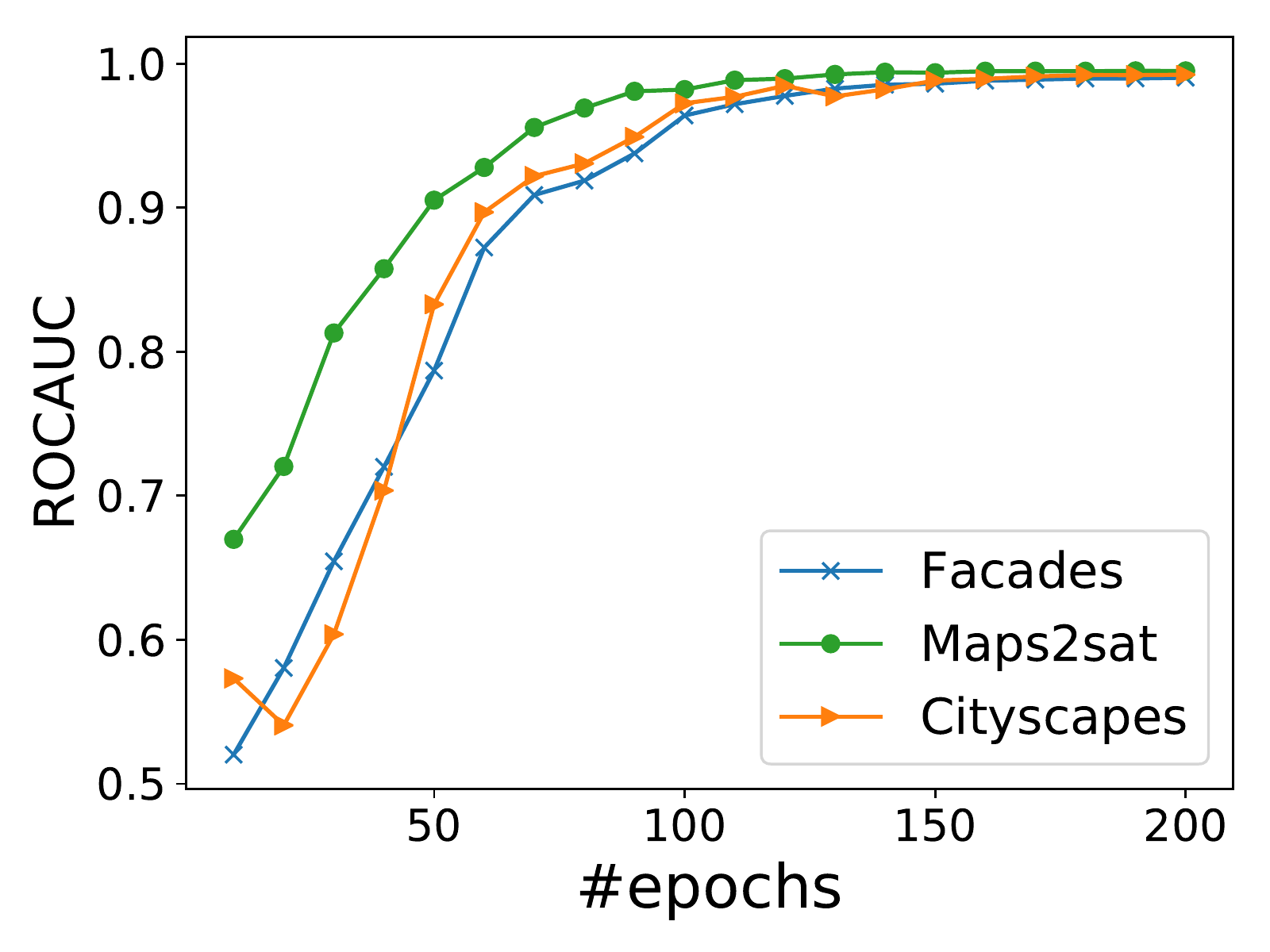}
}
\end{center}
\caption{Effect of memorization on the attack success rate.}
\label{fig:overfitting_effect}
\end{figure}

As mentioned in Sec.~\ref{sec:discussion}, memorization is the main reason for the success of our method. Fig.~\ref{fig:overfitting_effect} shows the accuracy of our method as a function of the number of epochs used for training the victim model, clearly suggesting that memorization is indeed the vulnerability.

\subsection{Defenses}
\begin{figure}[h]
\begin{center}
\subfigure[Pix2pix]{
\includegraphics[width=0.4\linewidth, height=0.35\linewidth]{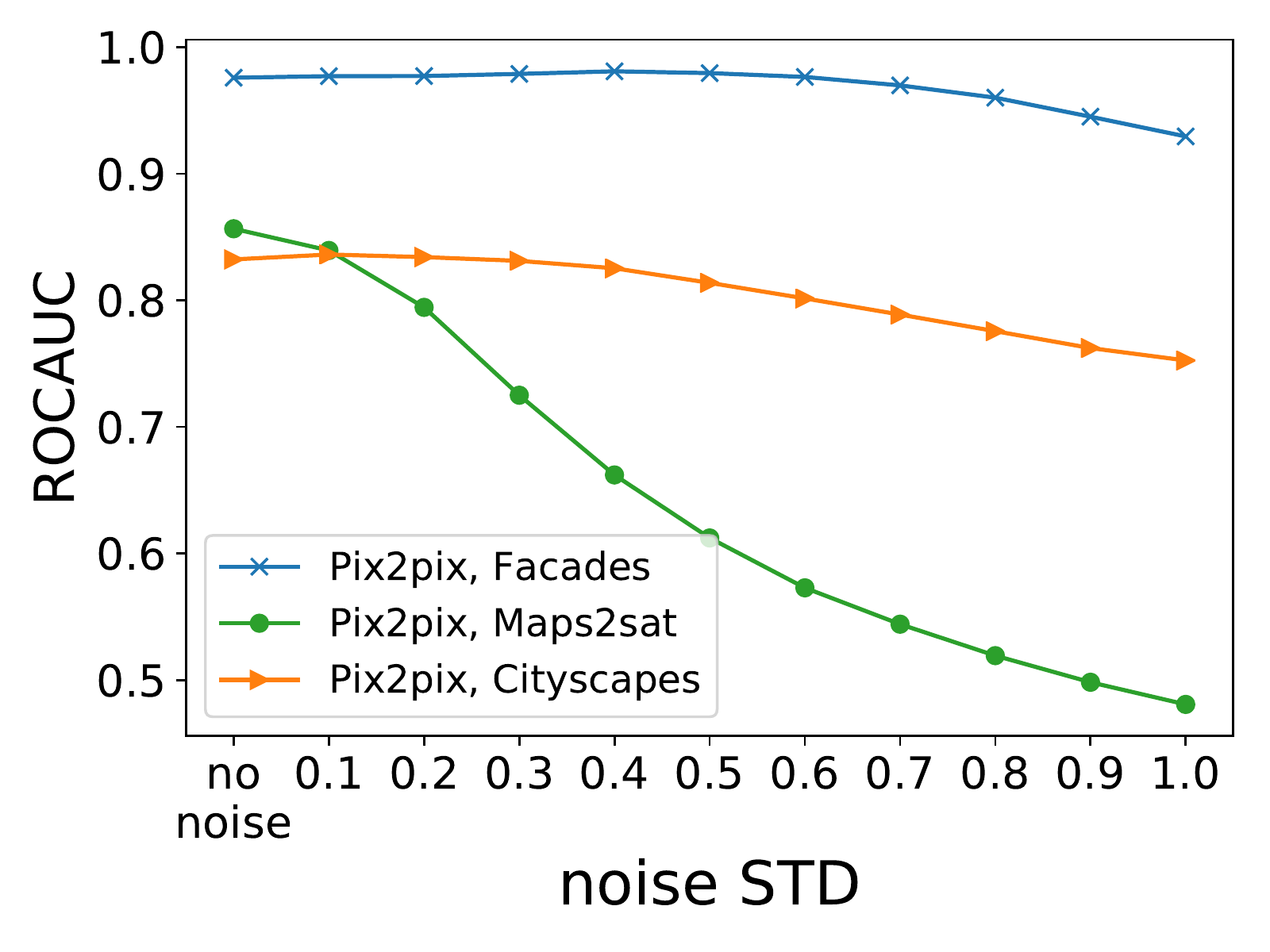}
}
\subfigure[SPADE]{
\includegraphics[width=0.4\linewidth, height=0.35\linewidth]{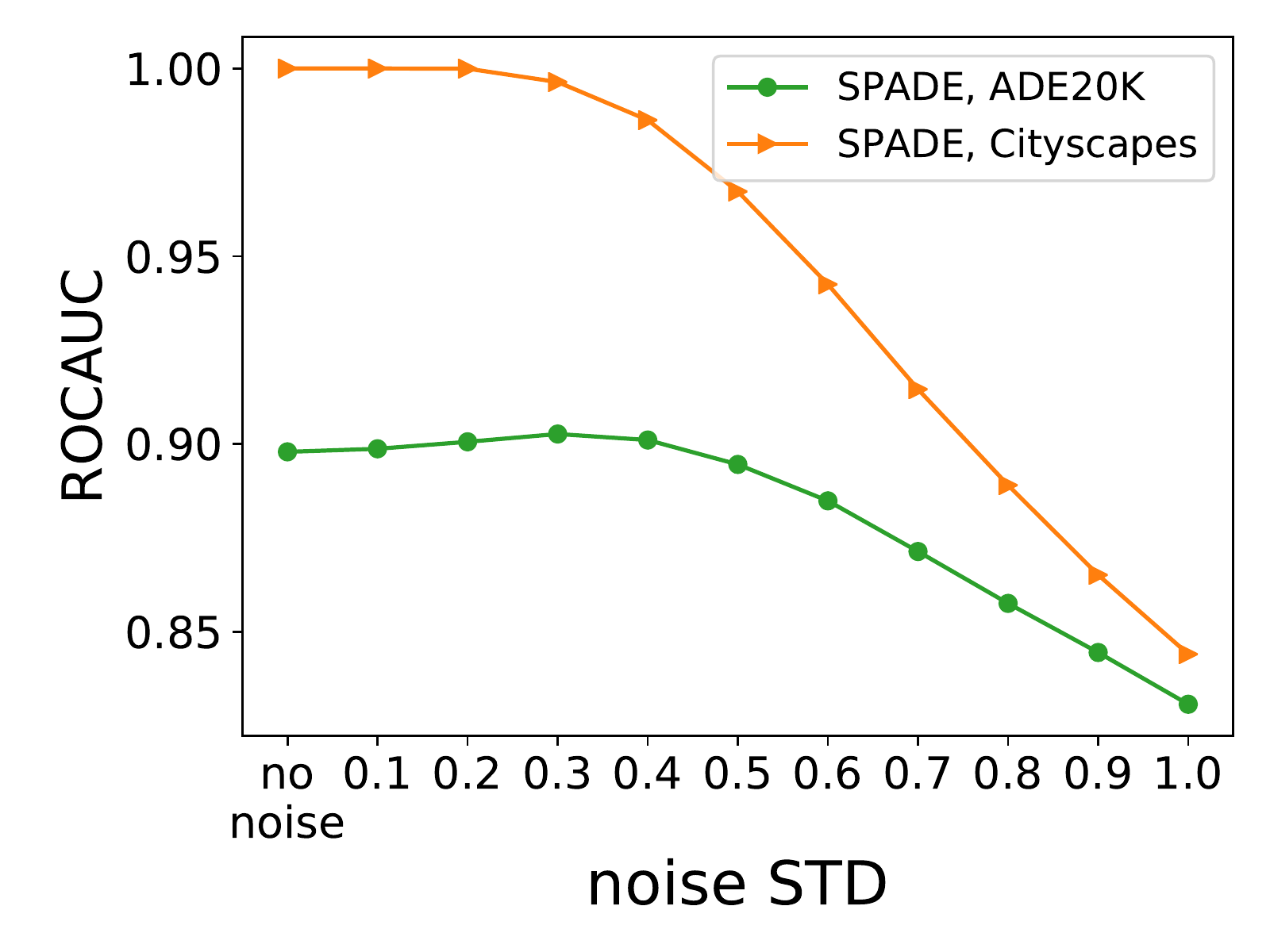}
}
\end{center}
\caption{Effect of Gauss defense on the attack success rate. Even with large amounts of added noise, our attack still manages to success much better then random guessing.}
\label{fig:rest_gauss_defense}
\end{figure}

In Sec.~\ref{sec:discussion}, we discuss the Gauss defense, including other common defenses, against our attack. We evaluated our attack accuracy as a function of different noise STD. Fig.~\ref{fig:rest_gauss_defense} shows that a considerable amount of noise, which corrupts the generated output, is required in order to have a significant effect over our attack success, which is still much better than random guessing ($50\%$). Results for Pix2PixHD, UperNet and HRNetV2 are presented in Fig.~\ref{fig:gauss_defense}.

\subsection{ImageNet predictability error}

\begin{figure}[h]
\begin{center}
\includegraphics[width=0.15\linewidth, height=0.18\linewidth]{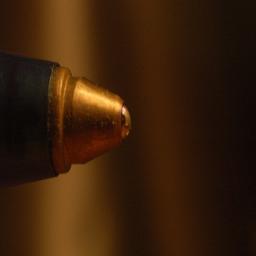}
\includegraphics[width=0.15\linewidth, height=0.18\linewidth]{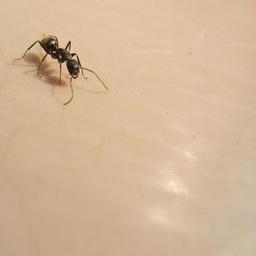}
\includegraphics[width=0.15\linewidth, height=0.18\linewidth]{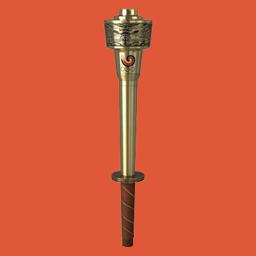}
\includegraphics[width=0.15\linewidth, height=0.18\linewidth]{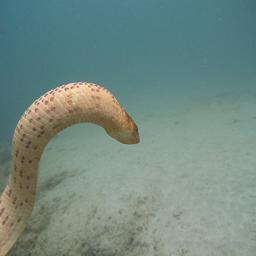}
\includegraphics[width=0.15\linewidth, height=0.18\linewidth]{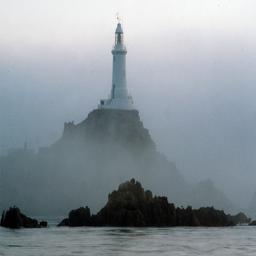}\\

\includegraphics[width=0.15\linewidth, height=0.18\linewidth]{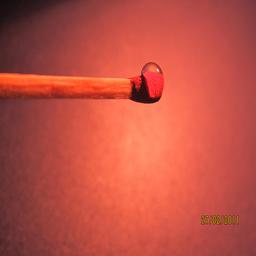}
\includegraphics[width=0.15\linewidth, height=0.18\linewidth]{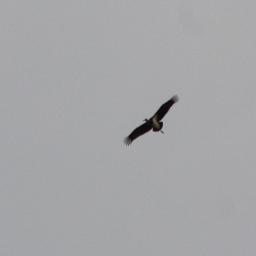}
\includegraphics[width=0.15\linewidth, height=0.18\linewidth]{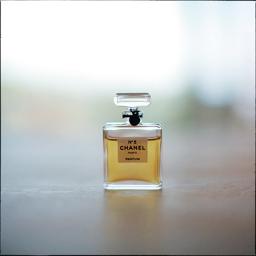}
\includegraphics[width=0.15\linewidth, height=0.18\linewidth]{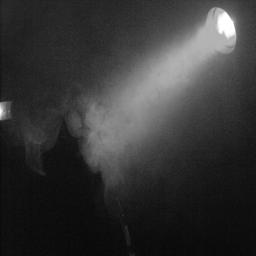}
\includegraphics[width=0.15\linewidth, height=0.18\linewidth]{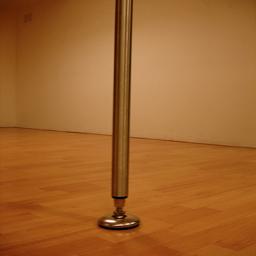}\\
\includegraphics[width=0.15\linewidth, height=0.18\linewidth]{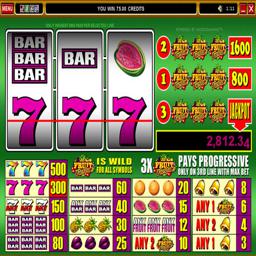}
\includegraphics[width=0.15\linewidth, height=0.18\linewidth]{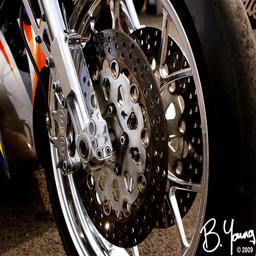}
\includegraphics[width=0.15\linewidth, height=0.18\linewidth]{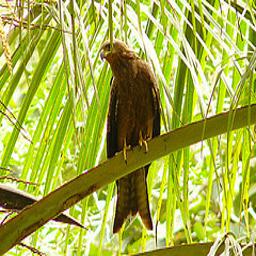}
\includegraphics[width=0.15\linewidth, height=0.18\linewidth]{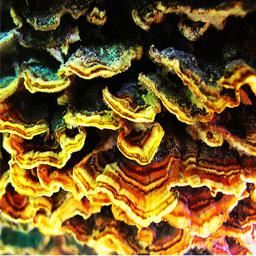}
\includegraphics[width=0.15\linewidth, height=0.18\linewidth]{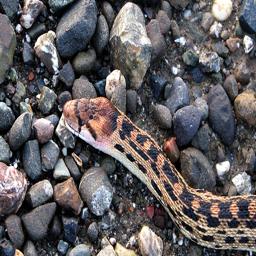}\\

\includegraphics[width=0.15\linewidth, height=0.18\linewidth]{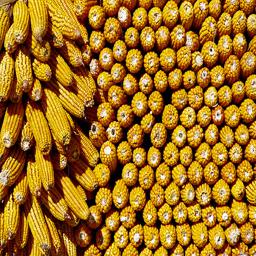}
\includegraphics[width=0.15\linewidth, height=0.18\linewidth]{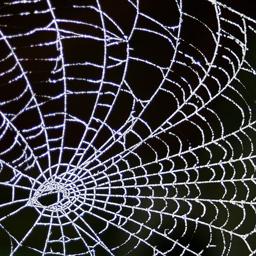}
\includegraphics[width=0.15\linewidth, height=0.18\linewidth]{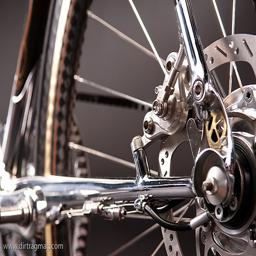}
\includegraphics[width=0.15\linewidth, height=0.18\linewidth]{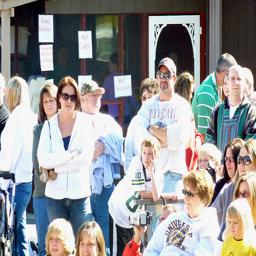}
\includegraphics[width=0.15\linewidth, height=0.18\linewidth]{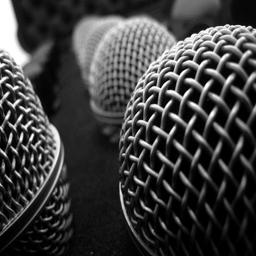}\\

\end{center}
\caption{Examples of images from the ImageNet dataset that received the lowest and highest predictability errors. First row - lowest scored train images. Second row - lowest scored test images. Third row - highest scored train images. Last row - highest scored test images. As can be seen, the predictability error is effective even on images that were used for training the feature extractor. }
\label{fig:imagenet_samples}

\end{figure}
Our predictability error relies on learning a mapping between feature vectors to their corresponding pixel values. We use a pre-trained Wide-ResNet$50{\times}2$ \cite{zagoruyko2016wide}, which is trained on the ImageNet dataset. We do not make any assumptions regarding an overlap between the pre-trained model's training data (i.e.\, ImageNet) and the data during in the attack. In the scenario in which such an overlap exists, the concern is that the predictability error would lose its credibility. 

In order to verify this, we computed the predictability error of a random subset of $1$K train images and $1$K test images, from the ImageNet dataset. As no input-output pairs exists, we trained the linear predictor to predict pixel values from deep features of the same image. We do not observe any significant difference between the two - both share similar mean and std values: ($0.0549$, $0.018$) for the train images and ($0.0556$, $0.0191$) for the test images. A ROCAUC score of $51\%$ further demonstrates that there is no clear difference between the distribution of the predictability error on seen and unseen images.

Fig.~\ref{fig:imagenet_samples} further demonstrates this. The first row presents the train images that received the lowest scores, i.e.\ marked as easy images, and the second row presents the test images with the lowest scores. Both correspond to "plain" images, regardless of whether they are known (train) or unknown (test). The same applies to the Difficult images. The third row presents the highest scored train images and the last row presents the highest scored test images. Both contains complex patterns and high variance. This demonstrates that the predictability error is not affected by the having prior knowledge of the image, and is only measuring the amount of variance and complexity of an image.

\end{document}